\documentclass[]{taotian}
\usepackage{enumitem}
\usepackage[toc,page,header]{appendix}

\usepackage[utf8]{inputenc} 
\usepackage[T1]{fontenc} 
\usepackage{hyperref} 
\usepackage{url}
\usepackage{array} 
\usepackage{booktabs}
\usepackage{amsfonts} 
\usepackage{nicefrac}
\usepackage{microtype} 
\usepackage{xcolor} 
\usepackage{xspace}
\usepackage{bm}
\usepackage{bbm}
\usepackage{bbding}
\usepackage{tabularx}
\usepackage{textcomp}
\usepackage{amssymb}
\usepackage{enumitem}
\usepackage{amsmath}
\usepackage{mathtools}
\usepackage{amsthm}
\usepackage{multirow}
\usepackage{makecell}
\usepackage{color}
\usepackage{colortbl}
\usepackage{adjustbox}
\usepackage{caption}
\usepackage{graphicx}
\usepackage{wrapfig}
\usepackage{array}
\usepackage{multicol}
\usepackage{lipsum}

\usepackage{algorithm}
\usepackage{algpseudocode}

\definecolor{myyellow}{RGB}{255,192,0}
\definecolor{mygreen}{RGB}{107,170,64}
\definecolor{mywrite}{RGB}{255,227,132}

\definecolor{policycolor}{HTML}{13501B}
\colorlet{policybgcolor}{policycolor!10}

\definecolor{worldcolor}{HTML}{C04F15}
\colorlet{worldbgcolor}{worldcolor!10}

\definecolor{bluearrow}{HTML}{156082}
\definecolor{purplearrow}{HTML}{78206E}

\definecolor{promptBlueBack}{HTML}{EBF5FF}
\definecolor{promptBlueFrame}{HTML}{003366}
\definecolor{promptGreenBack}{HTML}{F0FFF0}
\definecolor{promptGreenFrame}{HTML}{004D00}
\definecolor{promptOrangeBack}{HTML}{FFF5E6}
\definecolor{promptOrangeFrame}{HTML}{D35400} 
\definecolor{mylightgray}{gray}{0.9}

\newcommand\pooltoollogo{\raisebox{-0.5ex}{\includegraphics[height=3ex]{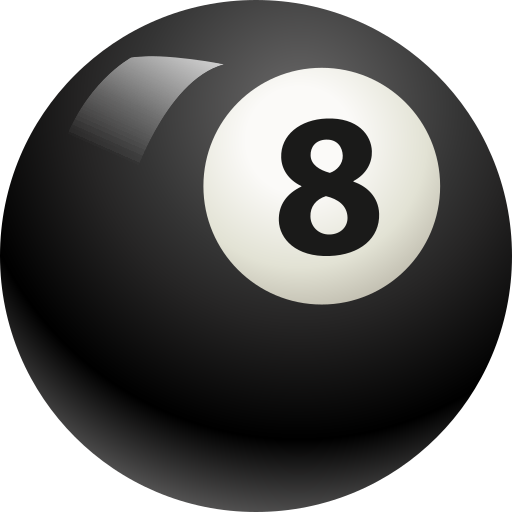}} \textbf{Pooltool}}
\newcommand\angrybirdlogo{\raisebox{-0.5ex}{\includegraphics[height=3ex]{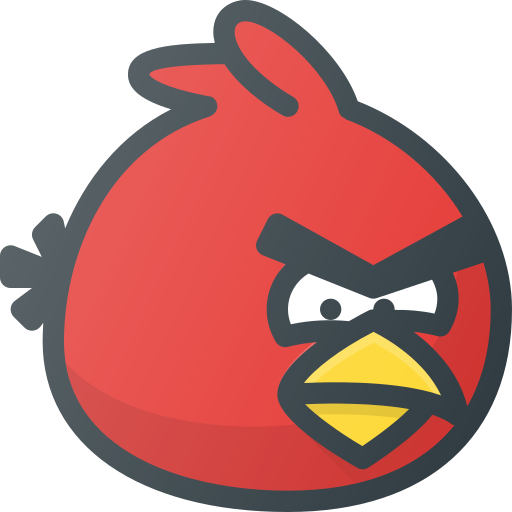}} \textbf{Angry Birds}}
\newcommand\cutrepologo{\raisebox{-0.5ex}{\includegraphics[height=3ex]{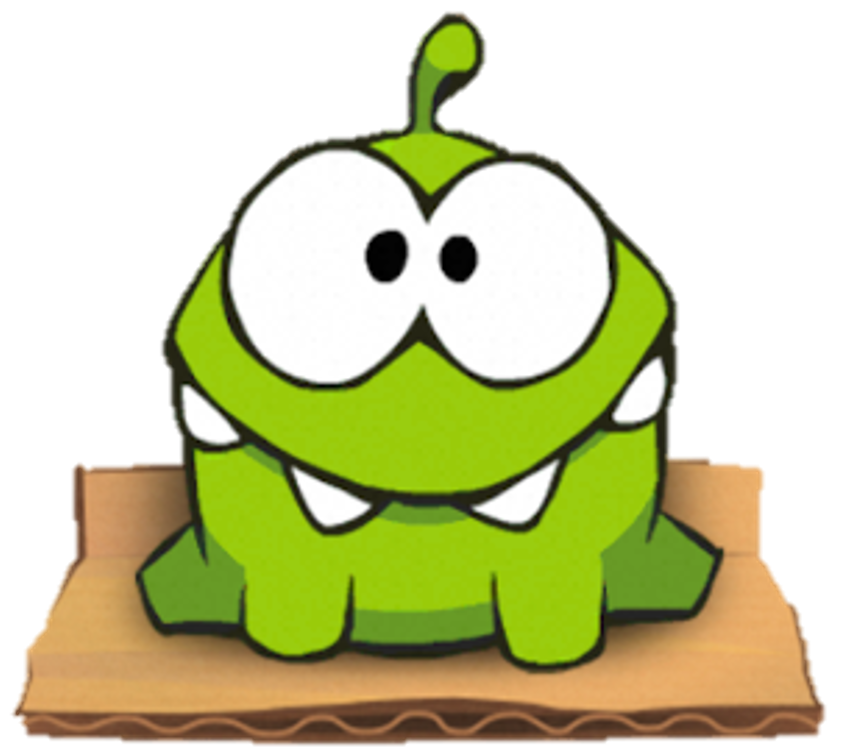}} \textbf{Cut the Rope}}
\newcommand\firstlogo{\raisebox{-0.5ex}{\includegraphics[height=2.5ex]{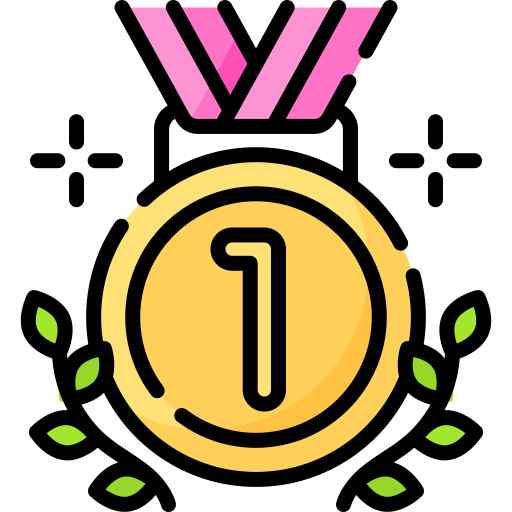}}}
\newcommand\secondlogo{\raisebox{-0.5ex}{\includegraphics[height=2.5ex]{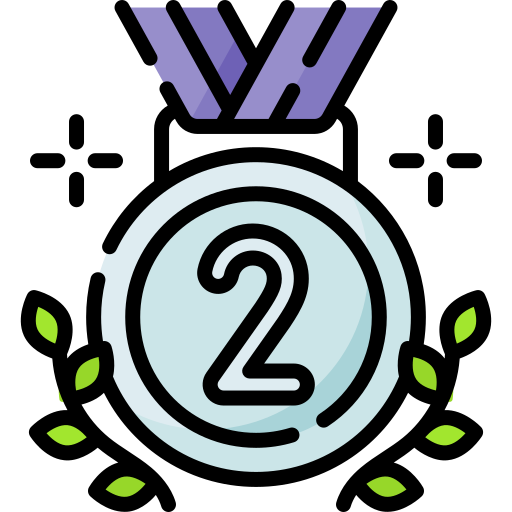}}}

\newcommand{\eg}{\textit{e.g., }}
\newcommand{\cf}{\textit{cf. }}

\newcommand{\LCommentWLN}[1]{\State \textbf{\textit{\small $\rhd$ #1}}}

\newcommand{\ProjName}{{ICPRL}\xspace}

\title{
    \begin{minipage}[c]{0.15\textwidth}
    \end{minipage}
    \begin{minipage}[r]{0.15\textwidth}
        \flushright
        \includegraphics[height=2cm]{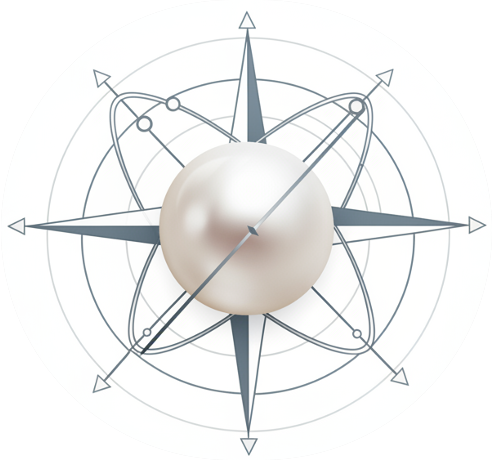}
    \end{minipage}\hfill
    \begin{minipage}[l]{0.65\textwidth}
        \centering
        \textbf{\ProjName:} Acquiring Physical Intuition\\
        from Interactive Control
    \end{minipage}\hfill
    \begin{minipage}[c]{0.1\textwidth}
    \end{minipage}
}

\author{{\bfseries 
Xinrun Xu$^{1,2}$, Pi Bu$^{3}$, Ye Wang$^{4}$, Börje F. Karlsson$^{5}$,\\Ziming Wang$^{3}$, Tengtao Song$^{3}$, Qi Zhu$^{3}$,\\Jun Song$^{3,*}$, Shuo Zhang$^{1,*}$, Zhiming Ding$^{1,*}$, Bo Zheng$^{3}$}}

\affiliation{
$^{1}$Institute of Software, Chinese Academy of Science,\\
$^{2}$University of Chinese Academy of Sciences,\\
$^{3}$Alibaba Group,
$^{4}$RUC,
$^{5}$PUC-Rio
}

\firstfig[width=\linewidth][\textwidth]{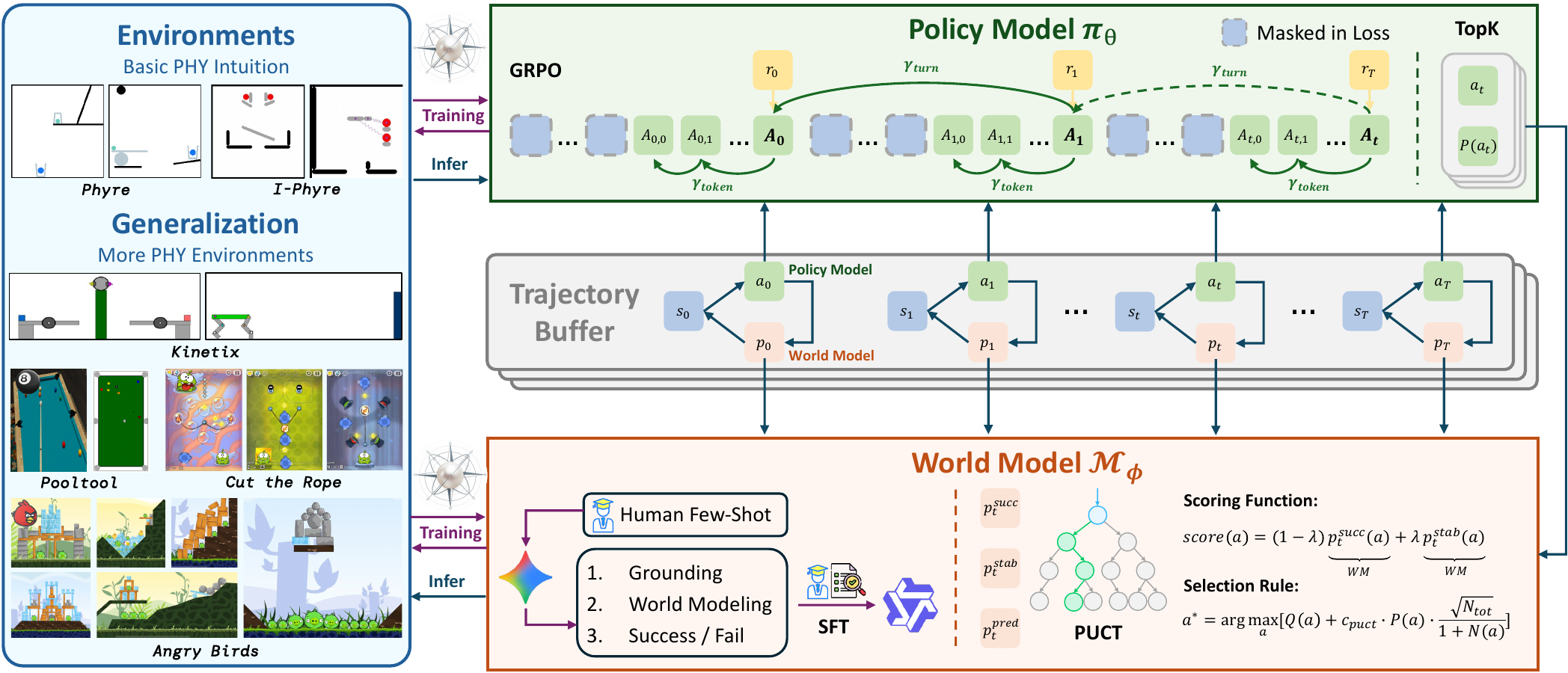}
{Overview of the \textbf{\ProjName} Framework, which decouples policy learning from world modeling for robust in-context planning. \textcolor{purplearrow}{\textbf{Training Stage:}} We separately train a \textcolor{policycolor}{\textbf{Policy Model ($\pi_\theta$)}} via \textbf{Turn-Aware GRPO} to generate context-aware actions---leveraging $\gamma_{\text{turn}}$ and $\gamma_{\text{token}}$ for precise multi-turn credit assignment---and a \textcolor{worldcolor}{\textbf{World Model ($\mathcal{M}_\phi$)}} to predict physical outcomes. \textcolor{bluearrow}{\textbf{Inference Stage:}} \textcolor{policycolor}{$\pi_\theta$} proposes candidate actions, which \textcolor{worldcolor}{$\mathcal{M}_\phi$} evaluates by acting as an \textit{in-context physical simulator}. These evaluations guide a PUCT search to select the optimal action, enabling effective zero-shot planning.}
{fig:icprl_framework}

\abstract{
VLMs excel at static perception but falter in interactive reasoning in dynamic physical environments, which demands planning and adaptation to dynamic outcomes. 
Existing physical reasoning methods often depend on abstract symbolic inputs or lack the ability to learn and adapt from direct, pixel-based visual interaction in novel scenarios. 
We introduce \textbf{\ProjName}\footnote{Pronounced \textit{IC-Pearl}.} (In-Context Physical Reinforcement Learning), a framework inspired by In-Context Reinforcement Learning (ICRL) that empowers VLMs to acquire physical intuition and adapt their policies in-context. 
Our approach trains a vision-grounded policy model via \textbf{multi-turn} Group Relative Policy Optimization (GRPO) over diverse multi-episode interaction histories. This enables the agent to adapt strategies by conditioning on past trial-and-error sequences, without requiring any weight updates. 
This adaptive policy works in concert with a separately trained world model that provides explicit physical reasoning by predicting the results of potential actions. 
At inference, the policy proposes candidate actions, while the world model predicts outcomes to guide a root-node PUCT search to select the most promising action.
Evaluated on the diverse physics-based puzzle-solving tasks in the DeepPHY benchmark, \ProjName demonstrates significant improvements across both its: \textbf{I.} policy-only, and \textbf{II.} world-model-augmented stages. 
Notably, these gains are retained in unseen physical environments, demonstrating that our framework facilitates genuine in-context acquisition of the environment's physical dynamics from interactive experience.

}

\definecolor{BlockC}{gray}{0.98}  
\definecolor{BlockA}{RGB}{191,211,230}
\definecolor{BlockB}{RGB}{199,233,192} 

\begingroup
 
\footnotetext[1]{Corresponding Authors.}
\endgroup

\begin{document}

\maketitle

\section{Introduction}

\begin{wrapfigure}{R}{0.45\linewidth}
\centering
\includegraphics[width=\linewidth]{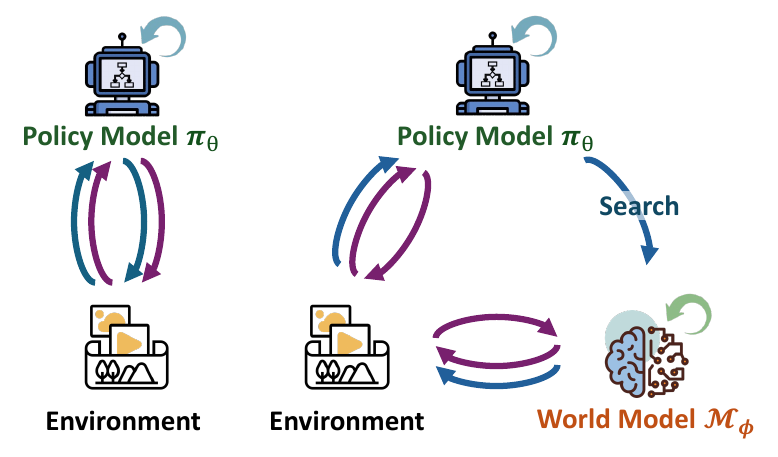}
\caption{{The \ProjName framework integrates world model and adaptive policy.} }
\label{fig:compare}
\end{wrapfigure}

Despite impressive perception, Vision-Language Models (VLMs) struggle with interactive physical reasoning that requires acting, observing consequences, and revising plans. Existing benchmarks largely test static understanding, leaving a gap in closed-loop, physics-grounded competence \citep{wang2023newtonlargelanguagemodels, agarwal2025cosmos}.
Prior work either distills policy improvement from symbolic histories, scales meta-RL \citep{duan2016rl} with test-time weight updates, or focuses on state-based off-policy in-context RL \citep{grigsby2023amago}; however, none directly equips VLMs to learn physics from raw pixels through interaction, to adapt to new tasks in a zero-shot setting \citep{keplinger2025ns}.

We introduce \textbf{\ProjName} (In-Context Physical Reinforcement Learning), an ICRL \citep{Bauer2023Ada} -inspired VLM framework that enables in-context policy adaptation for interactive physical reasoning. 
Our approach reinterprets the concept of distilling a policy improvement operator. Instead of training a model to \textit{imitate} the learning trajectory of a separate RL algorithm, our vision-grounded \textcolor{policycolor}{policy model ($\pi_{\theta}$)} directly becomes the subject of policy improvement. 
We train this VLM using \textbf{Turn-Aware GRPO} \citep{shao2024grpo}---a multi-turn adaptation of Group Relative Policy Optimization designed for precise temporal credit assignment---over a vast collection of diverse, multi-episode interaction histories. 
This process teaches the model \textit{how} to adjust its strategy based on historical context, including previous successes and failures. Consequently, the ability to adapt to new unseen task instances is not an emergent phenomenon at test time, but rather a robust capability learned and encoded into the model's weights during training. At inference, $\pi_{\theta}$ leverages this ability to improve its performance purely by conditioning on recent interaction history within its forward pass, requiring no further gradient updates.
Crucially, we augment this adaptive policy with an independently trained \textcolor{worldcolor}{world model ($M_{\phi}$)}, which acquires and provides explicit physical intuition by predicting action outcomes and dynamics. This world model serves as a powerful in-context planning component, guiding the policy's exploration and enabling the agent to discover task-specific physics and refine its plans more efficiently \citep{zhang2025abenchphysics, chung2025theoreticalphysics}.

This two-component architecture is illustrated in Figure~\ref{fig:compare}. Our \ProjName agent (\textbf{Right}) consists of: (1) an adaptive \textbf{\textcolor{policycolor}{Policy Model}} that learns to improve from its interaction history, and (2) an independently trained \textbf{\textcolor{worldcolor}{World Model}} that provides physical intuition. Training interactions for our models are denoted by \textcolor{purplearrow}{\textbf{purple arrows}}. At inference time (\textcolor{bluearrow}{\textbf{blue arrows}}), the policy performs an in-context {Search} over plans simulated by the world model, enabling more efficient reasoning. This approach stands in contrast to conventional agents (\textbf{Left}), which train a policy to react directly to environmental feedback without an explicit planning module. 

We evaluate \ProjName on DeepPHY \citep{xu2025deepphy}, a physical benchmark suite spanning single-shot placement to multi-step control, with standardized annotated observations and structured action spaces that preserve physics while enabling reliable VLM control \citep{wu2024ivideogpt}.
On the complex multi-step \textit{I-PHYRE} task, our full model achieves a {93.3\%} success rate. More critically, \ProjName shows remarkable generalization to unseen environments; \eg on the \textit{Pooltool} task, it achieves a {71.0\%} success rate, more than doubling the performance of strong baselines like GPT-o3. These gains on tasks the model was never trained on demonstrate that our framework facilitates genuine in-context acquisition of the environment’s physical dynamics from interactive experience.

Building on this evidence, our contributions are twofold: First, we introduce a novel vision-grounded ICRL-inspired framework for VLMs that adapts at test time without weight updates, enabled by a {Turn-Aware GRPO} mechanism that effectively distills physical intuition from interaction histories. Second, we establish state-of-the-art zero-shot performance on diverse interactive physical reasoning tasks, along with analyses on world-model prediction, action discretization, and history length.

\section{Related Works}

\subsection{In-Context Reinforcement Learning (ICRL)}

ICRL represents a paradigm shift in how autonomous agents acquire new skills, moving from slow, gradient-based adaptation to rapid ICL within the forward pass of neural networks. Typically, Algorithm Distillation (AD) \citep{Michael2023ICRL} distills a policy improvement operator into a Transformer by training on numerous learning histories from a source RL algorithm (\eg A3C \citep{Volodymyr2016}). 
While methods like AD focus on distilling an external RL algorithm—training a Transformer to \textit{imitate} the learning process itself—our \textbf{\ProjName} internalizes this capability directly within the policy model. 
To circumvent high computational cost, AD$^{\epsilon}$ \citep{Ilya2024ADepsilon} creates synthetic learning histories from a single demonstrator policy by simulating a learning process with a decaying noise curriculum. 
An alternative approach is memory-based meta-RL \citep{Bauer2023Ada}, in which an agent with persistent memory is trained to improve its strategy over repeated trials on the same task. 
Such techniques have been extended to LLMs by PAPRIKA \citep{tajwar2025training}, fine-tuning on self-generated interaction data using RPO \citep{Richard2024RPO}.
SCoRe \citep{Aviral2025SCoRe} uses on-policy RL to fine-tune an LLM, teaching it a generalizable ICL skill for multi-turn self-correction. ICAL \citep{Gabriel2024ICRA} leverages a VLM to autonomously refine noisy interaction data into high-quality examples, enhancing retrieval-augmented ICL planning.

The challenge of credit assignment in multi-turn LLM interactions has attracted growing attention. ArCHer~\citep{zhou2024archer} proposes an off-policy actor-critic framework that assigns per-turn credit through a hierarchical value function, enabling stable RL training over long interaction horizons. DAPO~\citep{yu2025dapo} addresses reward hacking and gradient instability in long-chain policy optimization at scale. Most directly relevant to our work, VAGEN~\citep{wang2025vagen} introduces selective token masking and turn-level advantage estimation for multi-turn VLM agent training---mechanisms we adopt and extend in our Turn-Aware GRPO. Crucially, none of these works targets the generalization of a learned improvement operator across \emph{physically diverse} environments; they are designed for single-task or domain-specific settings, whereas \ProjName is explicitly trained to transfer physical intuition zero-shot to unseen domains.

ICRL can also leverage Unsupervised Environment Design (UED) and automated curriculum learning to generate diverse and high-quality training tasks.
Building on XLand \citep{openendedlearningteam2021openended}, which advanced the training of generalist agents through an open-ended process of dynamic task generation, Adaptive Agent (AdA) \citep{Bauer2023Ada} introduces an automated curriculum method to navigate the immense XLand 2.0 task space.
Prioritized Level Replay (PLR) \citep{JiangGR21} is a heuristic that focuses training on tasks at the edge of an agent's competence, and has been theoretically formalized and improved \citep{JVBRCL20, JiangDPFGR21, ParkerHolderJ022}.
Recent curriculum strategies have advanced from refining task selection at the learning frontier \citep{RutherfordBWLHF24} to augmenting the training space with generative models \citep{GarcinDGLA24}, and toward creating truly open-ended domains by evolving the game mechanics themselves \citep{earle2024autoverseevolvablegamelanguage, FransI23}.

Instead of mimicking an optimization algorithm, our VLM policy is trained via Turn-Aware GRPO\citep{shao2024grpo} on multi-episode histories to learn \textit{how to execute} a policy improvement step in-context. 
By observing diverse histories of trial and error, the VLM learns to condition its future actions on past outcomes, effectively embedding an adaptive, in-context learning skill into its own parameters. 
This is achieved through gradient-based optimization over multi-episode histories, enabling the VLM to perform in-context adaptation at inference time.

\subsection{Physical Reasoning in VLMs}

Physical reasoning serves as the foundation for world model construction \citep{wu2024ivideogpt,agarwal2025cosmos} and embodied intelligence tasks \citep{luo2023model,lin2025switch,Being-0}. However, most evaluations of LLMs and VLMs focus on static problem-solving benchmarks. These evaluations—often large-scale QA tasks on object properties \citep{wang2023newtonlargelanguagemodels,chow2025physbench} or text-based physics exams \citep{wang2025phy,chung2025theoreticalphysics,zhang2025abenchphysics}—assess primarily the agents' ability to recall scientific knowledge or infer logical outcomes from fixed contexts. While useful for evaluating declarative knowledge, these approaches largely bypass the challenges of real-time visual perception and continuous interaction in dynamic environments.

To evaluate interactive physical reasoning, another line of research investigates agents in simulated environments. However, these works often abstract away perceptual grounding by relying on pre-processed symbolic inputs, such as object property matrices \citep{bakhtin2019phyre,iphyre,Kinetix}, or by enabling interaction via code generation \citep{LLMPhy}. Within benchmarks like PHYRE specifically, specialized RL methods have achieved substantially higher performance than general-purpose models by leveraging Deep Q-Learning, relational neural physics models, and improved exploration strategies. While effective for isolating specific planning tasks, these approaches are tightly coupled to the specific simulator, action space, and state representation, providing no path to zero-shot generalization and bypassing raw sensory data understanding. Similarly, while digital game agents \citep{tan2024cradle, chen2025combatvla} process raw observations, they typically require only a shallow understanding of common-sense physics.

A parallel line of research employs learned world models to enable planning in physical environments through model-based RL. For instance, DreamerV3~\citep{hafner2023dreamerv3} learns a compact latent-space world model to optimize behavior entirely within imagination, while TD-MPC2~\citep{hansen2024tdmpc2} combines model predictive control with learned temporal difference models for continuous control. While these methods demonstrate the power of world models for physical reasoning, they operate on low-level continuous state representations, require environment-specific training from scratch, and provide no mechanism for zero-shot generalization to novel tasks from raw visual observations.

Our work addresses these limitations by focusing on interactive physical reasoning directly from visual inputs, where agents must plan and execute a sequence of actions guided by learned intuition. \ProjName trains a world model on only two source environments and applies it zero-shot as an in-context physical simulator across distinct target domains. Unlike recent candidate-filtering approaches \citep{qi2025strengthening}—where a world model serves merely as a one-shot, static discriminator to score a fixed set of actions in a single pass—\ProjName enables genuine test-time adaptation through a world-model-guided lookahead search. Specifically, \ProjName's PUCT search \emph{iteratively} queries the world model over $B$ planning iterations, maintaining per-action visit counts $N(a)$ and value estimates $Q(a)$. This progressive search dynamically balances the exploitation of high-scoring candidates with the exploration of less-visited alternatives. Furthermore, \ProjName trains the policy to internalize this improvement operator via Turn-Aware GRPO, removing the reliance on task-specific fine-tuning. We evaluate our approach on DeepPHY \citep{xu2025deepphy}, covering six diverse dynamic physical environments, to demonstrate \ProjName's superior interactive physical reasoning capabilities.

\section{\ProjName}

We consider a family of physical simulators $\{\mathcal{E}_m\}_{m=1}^M$. At each decision point we observe $s_t\in\mathcal{S}$, choose an action $a_t\in\mathcal{A}$, and the simulator returns an updated observation and a task reward or success indicator. Our \ProjName framework comprises distinct training and inference stages, as shown in Figure \ref{fig:icprl_framework}:
\textbf{\textcolor{purplearrow}{Training (two stages):}} We foster \textbf{in-context adaptive capabilities} within a VLM \textbf{\textcolor{policycolor}{policy model $\pi_\theta$}} through online GRPO \citep{shao2024grpo} on diverse multi-episode interaction histories. Concurrently, a separate \textbf{\textcolor{worldcolor}{world model $\mathcal{M}_\phi$}} is trained offline to acquire robust physical intuition by predicting environment dynamics and outcomes.
\textbf{\textcolor{bluearrow}{Inference:}} These two models work in concert. $\pi_\theta$ proposes contextually informed actions, and $\mathcal{M}_\phi$, acting as an \textbf{in-context physical simulator}, provides crucial outcome predictions to guide a lightweight \textbf{root-node PUCT} \citep{silver2017mastering} search. This separation keeps policy $\pi_\theta$ learning stable and simulator-grounded, while enabling rapid context-sensitive adaptation and refined planning at test time without further weight updates.

\subsection{Mathematical Definitions}
\label{app:mathematical_definitions}

In this section, we provide the formal definitions for the components used in the \ProjName framework to ensure reproducibility and clarity.

We formalize the interactive physical reasoning task as a multi-step Partially Observable Markov Decision Process (POMDP) \citep{POMDP} augmented with interaction history.

\begin{itemize}
    \item \textbf{State Space ($\mathcal{S}$):} The underlying physical state of the simulator at time $t$ (denoted as $s_t$), such as object positions, velocities, and friction coefficients. Note that $s_t$ is not directly accessible to the agent.

     \item \textbf{Observation Space ($\mathcal{O}$):} The visual rendering of the state $o_t \in \mathcal{O}$ at time step $t$. This may include annotated overlays as detailed in Section \ref{sec:experimental_setup} and DeepPHY \cite{xu2025deepphy}.

    \item \textbf{Action Space ($\mathcal{A}$):} The set of executable commands. While the physics simulator accepts continuous parameters, we discretize $\mathcal{A}$ into text tokens for VLM processing. For example, in Angry Birds, continuous angles $\theta \in [0, 90]$ are discretized into integer bins.

    \item \textbf{Terminal Reward ($\mathcal{R}$):} We utilize a sparse binary reward signal $r_{GT}$ provided by the environment at the end of an episode of length $T$. Specifically, $r_{GT}=1$ denotes success, while $r_{GT}=0$ denotes failure. No intermediate rewards are provided.

    \item \textbf{Trajectory ($\tau$):} A sequence recording the interaction data of a single attempt from start to finish: 
    \begin{equation*}
        \tau = (o_0, a_0, o_1, a_1, \dots, o_T, a_T, r_{GT})
    \end{equation*}

     \item \textbf{Interaction History ($H$):} A collection of trajectories from past failed attempts within the same problem instance: $H_k = (\tau_1, \tau_2, \dots, \tau_{k-1})$.

    \item \textbf{Policy Mapping:} The policy $\pi_\theta(a_t \mid o_{\le t}, H)$ maps the current observation sequence (context) and past interaction history to a probability distribution over the next action $a_t$.

    \item \textbf{Reference Policy ($\pi_{ref}$):} A frozen version of the policy model derived from SFT stage. It serves as the anchor for the KL-divergence penalty, preventing the RL-tuned policy $\pi_\theta$ from deviating excessively from the natural language distribution and maintaining output format validity.

\end{itemize}

\subsection{Training Stage: Policy Model $\pi_\theta$}

Our \textcolor{policycolor}{policy model $\pi_\theta$} is a VLM that generates textual outputs in a structured format (detailed in Section \ref{subsec:Inference}). These outputs are parsed into discrete actions $a_t \in \mathcal{A}$ by an environment-specific converter, for interaction with simulators across $M$ environments. The reinforcement signal is derived directly from the task's native rewards (\eg success/failure).

While standard RL algorithms can be applied to this setup by treating the agent's trajectory as a monolithic sequence, this approach often proves suboptimal. It fails to distinguish between agent-generated tokens (reasoning and actions) and environment-provided context, and it applies a uniform temporal discounting that conflates intra-turn and inter-turn credit assignment. To address these limitations, we enhance our Group Relative Policy Optimization (GRPO) framework by incorporating principles from selective token masking and multi-turn credit assignment, inspired by VAGEN framework\cite{wang2025vagen}. 

First, we employ {selective token masking} to focus the learning signal strictly on the agent's decision-making process. We introduce a mask $M^{\text{loss}}_t \in \{0, 1\}$, where $M^{\text{loss}}_t = 1$ for all tokens generated by the policy $\pi_\theta$ (reasoning and action tokens) and $M^{\text{loss}}_t = 0$ for all observation and prompt tokens. Second, for multi-turn interaction histories, standard GRPO assigns a single sequence-level advantage, which leads to sparse and unstable gradients over long horizons. To resolve this, we adopt a {Turn-Aware Group Advantage Estimation} scheme. Instead of a single sequence-level advantage, we calculate a turn-specific advantage $A_{i,k}$ for the $k$-th turn of the $i$-th sampled trajectory within a group. This effectively implements an inter-turn discount $\gamma_{\text{turn}}$ for future state transitions, while treating all generated tokens within a single turn (intra-turn, $\gamma_{\text{token}}=1.0$) as equally responsible for that turn's outcome.

Formally, at a given decision step, we sample a group of $G$ output sequences from the policy model $\pi_\theta$, denoted as $\mathcal{G} = \{o_1, o_2, \dots, o_G\}$. The importance sampling ratio for a given token $t$ in the $i$-th sample is defined as $r_{i,t}(\theta) = \frac{\pi_{\theta}(o_{i,t} \mid q, o_{i,<t})}{\pi_{\theta_{\text{old}}}(o_{i,t} \mid q, o_{i,<t})}$.

Let $A_{i,k}$ denote the relative advantage for the $k$-th turn of the $i$-th trajectory, computed by standardizing the turn-specific returns within the group:
\begin{equation}
    A_{i,k} = \frac{R_{i,k} - \text{mean}(R_{*,k})}{\text{std}(R_{*,k})}
\end{equation}
where $R_{i,k}$ is the discounted return from turn $k$ onwards ($R_{i,k} = \sum_{l=k}^K \gamma_{\text{turn}}^{l-k} r_{i,l}$). 
For a trajectory with $K$ turns, the token-level GRPO objective function (to be maximized) is formulated as:

\begin{align}
    \mathcal{J}_{\text{GRPO}}(\theta) = \frac{1}{G} \sum_{i=1}^{G} \sum_{k=1}^{K} \frac{1}{\sum_t M^{\text{loss}}_{i,k,t}} \sum_{t \in \text{turn}_k} M^{\text{loss}}_{i,k,t} \Bigg[ & \min\left(r_{i,t}(\theta) \cdot A_{i,k}, \ \text{clip}\left(r_{i,t}(\theta), 1-\varepsilon, 1+\varepsilon\right) \cdot A_{i,k}\right) \nonumber \\
    & - \beta D_{\text{KL}}\left( \pi_{\theta}(\cdot \mid q, o_{i,<t}) \,\|\, \pi_{\text{ref}}(\cdot \mid q, o_{i,<t}) \right) \Bigg]
\end{align}

Here, $M^{\text{loss}}_{i,k,t}$ indicates whether the $t$-th token in turn $k$ is generated by the policy, $\varepsilon$ is the clipping hyperparameter, and $\beta$ is the coefficient for the Kullback-Leibler (KL) divergence penalty against the frozen reference model $\pi_{\text{ref}}$. By coupling GRPO with multi-turn masking, we achieve precise token-level credit assignment without the computational overhead of training a separate Critic model.

\subsection{Training Stage: World Model $\mathcal{M}_\phi$}

Given an observation-action pair $(o,a)$, the \textcolor{worldcolor}{world model $\mathcal{M}_\phi$} is trained to return a strictly structured JSON containing a predicted success probability and a natural language prediction.

\begin{align}
    \mathcal{M}_\phi(o,a) \Rightarrow
    \begin{cases}
        \hat{p}_{\mathrm{succ}} \in [0,1], \\
        \hat{p}_{\mathrm{pred}}: \text{NL prediction}
    \end{cases}
\end{align}

The predicted success probability $\hat{p}_{\mathrm{succ}}$ is used directly for planning, while $\hat{p}_{\mathrm{pred}}$ provides qualitative insights for analysis. 

To enhance planning robustness, a stability score, $\hat{p}_{\mathrm{stab}}$, is derived from the model's own predictions. This score is calculated by evaluating the model's success predictions over a neighborhood of perturbed actions. For an environment-specific action metric $d_\mathcal{A}$ and a radius $\delta$, we define the perturbation set as:
$\mathcal{B}_\delta(a)=\{a': d_\mathcal{A}(a,a')\le\delta\}$.
The stability $\hat{p}_{\mathrm{stab}}$ is the expected model-predicted success rate over actions sampled from this perturbation set. Let $\hat{p}_{\mathrm{succ}}(o,a')$ denote the success probability predicted by $\mathcal{M}_\phi(o,a')$. The stability is then defined as:
$\hat{p}_{\mathrm{stab}}(o,a) = \mathbb{E}_{a'\sim D(\mathcal{B}_\delta(a))}[\,\hat{p}_{\mathrm{succ}}(o,a')\,]$.

In practice, this expectation is estimated for each sample by querying the model $\mathcal{M}_\phi$ with a small, finite set of ``jittered'' actions and averaging the resulting $\hat{p}_{\mathrm{succ}}$ predictions.

In practice, this loop terminates rapidly. In physical puzzle
environments, the vast majority of actions result in failure;
successful actions occupy only a small fraction of the action
space. Consequently, each random sample drawn from $\mathcal{A}$
is overwhelmingly likely to be a failing action, ensuring that
$k$ diverse failure samples are collected within a small,
bounded number of iterations across all evaluated environments.

\begin{algorithm}[htbp]
\caption{World Model Dataset Curation.}
\label{alg:wm_data_curation}
\begin{algorithmic}[1]
\Require Task set $\mathcal{X}$; action space $\mathcal{A}$; simulator $\textsc{Sim}(x,a)$ returns trajectory $\tau$ and label $y \in \{0,1\}$; frames $m{=}5$; diversity threshold $\varepsilon$.
\Ensure WM dataset $\mathcal{D}_{\mathrm{WM}}$.

\State $\mathcal{D}_{\mathrm{WM}} \gets \emptyset$

\For{each task $x \in \mathcal{X}$}
    \State Get initial visual states: $I_0, I_{\mathrm{ann}}$
    
    \State \Comment{\textbf{Phase 1: Solution Discovery}}
    \State $S \gets \{a \in \mathcal{A} \mid \textsc{Sim}(x,a).y = 1\}$ \Comment{Enumerate or retrieve cached solutions}
    \State $k \gets |S|$
    \If{$k=0$} \textbf{continue} \EndIf

    \State \Comment{\textbf{Phase 2: Balanced Failure Sampling}}
    \State $F \gets \emptyset$
    \While{$|F| < k$}
        \State Sample candidate $a \sim \mathcal{A}$
        \State $d_{\min} \gets \min_{a' \in S \cup F} \mathrm{dist}(a, a')$
        \If{$d_{\min} \ge \varepsilon$ \textbf{and} $\textsc{Sim}(x,a).y = 0$}
            \State $F \gets F \cup \{a\}$
        \EndIf
    \EndWhile

    \State \Comment{\textbf{Phase 3: VLM Annotation \& Compilation}}
    \For{each action $a \in S \cup F$}
        \State $(\tau, y) \gets \textsc{Sim}(x,a)$
        \State Extract $m$ uniformly spaced frames $F_{1:m}$ from $\tau$ post-action
        
        \State \Comment{Generate Grounding, Reasoning, and Label via VLM}
        \State $T_{\text{prompt}} \gets (I_0, I_{\mathrm{ann}}, a, F_{1:m}, y)$
        \State $T_{\text{output}} \gets \textsc{VLM}(T_{\text{prompt}})$
        
        \If{human verification passes} 
            \State $\mathcal{D}_{\mathrm{WM}} \gets \mathcal{D}_{\mathrm{WM}} \cup \{(x, I_0, I_{\mathrm{ann}}, a, F_{1:m}, y, T_{\text{output}})\}$ 
        \EndIf
    \EndFor
\EndFor
\State \Return $\mathcal{D}_{\mathrm{WM}}$
\end{algorithmic}
\end{algorithm}

To train world model $\mathcal{M}_\phi$, we first curate a high-quality dataset with a balanced representation of outcomes. This procedure is formalized in Algorithm~\ref{alg:wm_data_curation}. The data generation process for each initial state $s$ is as follows:

We first enumerate all possible successful actions from state $s$. If $k$ successful actions are found, we then sample and filter to obtain a corresponding set of $k$ diverse, failing actions. This ensures a balanced 1:1 ratio of positive to negative examples for each state, preventing model bias.
In practice, this loop terminates rapidly. In physical puzzle environments, the vast majority of actions result in failure; successful actions occupy only a small fraction of the action space. Consequently, each random sample drawn from $\mathcal{A}$ is overwhelmingly likely to be a failing action, ensuring that $k$ diverse failure samples are collected within a small, bounded number of iterations across all evaluated environments.

Each successful and failing action is executed in the simulator. We record the entire chain reaction following the action, from its completion until the environment reaches a stable state. From this resulting video sequence, we uniformly sample 5 frames to represent the dynamic evolution of the environment post-action.
The collected data—comprising the initial state (raw and annotated images), the action, the five dynamic frames, and the ground-truth success/fail label—is used to generate rich training signals. We employ a large VLM with a few-shot prompting strategy to produce the NLP ground truth for $\hat{p}_{\mathrm{pred}}$. 
The VLM is prompted to generate text describing the objects and their relations in the initial state (\textbf{Grounding}), predicting the chain reaction that will occur as a consequence of the action (\textbf{World Modeling}), and providing a \textbf{Success or Fail Label}. All automatically generated annotations are then manually inspected and corrected by human reviewers to ensure their accuracy and quality.

With this curated dataset, we train world model $\mathcal{M}_\phi$. The training objective combines a regression loss for success prediction and a language modeling loss for the textual output. The total loss $\mathcal{L}_{\mathrm{WM}}$ is a weighted sum:
$
    \mathcal{L}_{\mathrm{WM}} = \mathcal{L}_{\mathrm{succ}} + \lambda_{\mathrm{text}}\mathcal{L}_{\mathrm{text}}
$,
where $\mathcal{L}_{\mathrm{succ}} = \mathrm{BCE}(\hat{p}_{\mathrm{succ}}, y)$ is the Binary Cross-Entropy loss between the predicted success probability and the ground-truth label $y$, and $\mathcal{L}_{\mathrm{text}}$ is a standard cross-entropy loss for training the language prediction component.

\subsection{Inference Stage}
\label{subsec:Inference}

At inference time, for each decision point, we employ a root-node search procedure that integrates a learned policy prior with scores from our world model, $\mathcal{M}_\phi$. This process, detailed in Algorithm~\ref{alg:inference}, unfolds in four stages: candidate generation, scoring, search, and execution.

\begin{algorithm}[htbp]
\caption{Inference with Root-Node Search.}
\label{alg:inference}
\begin{algorithmic}[1]
\Require Observation $o$; policy $\pi_\theta$; world model $\mathcal{M}_\phi$; scoring strategy $\texttt{strat} \in \{1, 2\}$; hyperparameters $S, B, c_{\mathrm{PUCT}}$; Strategy~1: $J, \lambda_{\mathrm{PUCT}}$; Strategy~2: $K, \lambda_{\mathrm{LCB}}$.
\Ensure Best action $a^*$.

\LCommentWLN{Stage 1: Candidate Generation and Prior}
\State $A_{\text{samples}} \gets \emptyset$
\For{$i = 1 \to S$}
    \State Sample $a \sim \pi_\theta(\cdot \mid o)$ and add to $A_{\text{samples}}$
\EndFor
\State $A \gets \textsc{Unique}(A_{\text{samples}})$
\For{each action $a \in A$}
    \State $P(a) \gets \textsc{Count}(a, A_{\text{samples}}) / S$
\EndFor

\LCommentWLN{Stage 2 \& 3: PUCT Search Guided by WM Scores}
\State $N(a) \gets 0, Q(a) \gets 0$ for all $a \in A$
\State $a_{\mathrm{last}} \gets \text{null}, \ n_{\mathrm{same}} \gets 0$

\For{$t = 1 \to B$}
    \State $N_{\mathrm{tot}} \gets \sum_{b \in A} N(b)$
    \State $a_t \gets \arg\max_{a\in A}\left[Q(a)+c_{\mathrm{PUCT}}\cdot P(a)\cdot \frac{\sqrt{N_{\mathrm{tot}}}}{1+N(a)}\right]$
    
    \State \Comment{Early stopping check}
    \If{$a_t = a_{\mathrm{last}}$} $n_{\mathrm{same}} \gets n_{\mathrm{same}} + 1$ \Else\ $n_{\mathrm{same}} \gets 1$ \EndIf
    \If{$n_{\mathrm{same}} \ge 3$} \textbf{break} \EndIf
    \State $a_{\mathrm{last}} \gets a_t$

    \State \Comment{Compute Score}
    \If{$\texttt{strat} = 1$} \Comment{Strategy 1: Action Space Perturbation (PHYRE, I-PHYRE, Angry Birds)}
        \State $\hat{p}_{\mathrm{succ}} \gets \mathcal{M}_\phi(o, a_t)$
        \State Sample $J$ perturbed neighbors $\{a'_j\}_{j=1}^J \sim \mathcal{B}_\delta(a_t)$
        \State $\hat{p}_{\mathrm{stab}} \gets \frac{1}{J} \sum_{j=1}^J \mathcal{M}_\phi(o, a'_j)$
        \State $v \gets (1-\lambda_{\mathrm{PUCT}}) \cdot \hat{p}_{\mathrm{succ}} + \lambda_{\mathrm{PUCT}} \cdot \hat{p}_{\mathrm{stab}}$
    \Else \Comment{Strategy 2: Model Confidence / LCB (Kinetix, Pooltool, Cut the Rope)}
        \State Sample $K$ temperatures $\{\tau_j\}_{j=1}^K$ uniformly from $[0.1, 1.0]$
        \State $p^{(j)} \gets \mathcal{M}_\phi(o, a_t; \tau_j)$ for $j = 1, \dots, K$
        \State $\mu_p \gets \frac{1}{K}\sum_{j=1}^K p^{(j)}$, \quad $\sigma_p \gets \sqrt{\frac{1}{K}\sum_{j=1}^K (p^{(j)} - \mu_p)^2}$
        \State $v \gets \mu_p - \lambda_{\mathrm{LCB}} \cdot \sigma_p$
    \EndIf
    
    \State \Comment{Update search statistics}
    \State $Q(a_t) \gets \frac{Q(a_t) \cdot N(a_t) + v}{N(a_t) + 1}$
    \State $N(a_t) \gets N(a_t) + 1$
\EndFor

\LCommentWLN{Stage 4: Final Selection}
\State $a^* \gets \arg\max_{a \in A} Q(a)$
\State \Return $a^*$
\end{algorithmic}
\end{algorithm}

\paragraph{Candidate Generation.} First, to form a discrete set of candidate actions $A=\{a_i\}_{i=1}^S$, we draw $S$ samples from our policy network $\pi_\theta(\cdot\mid o)$. Based on these samples, we establish a \textbf{frequency prior} $P(a)$ over the candidate set, defined as $P(a) = c(a) / \sum_{a' \in A} c(a')$, where $c(a)$ is the count of $a$.

\paragraph{Scoring Strategy.} Next, each candidate action $a \in A$ is evaluated by $\mathcal{M}_\phi$. To handle the distinct characteristics of varying physical environments—ranging from stable to chaotic dynamics—we employ two tailored scoring strategies, summarized in Table~\ref{tab:stability_computation}.

\begin{table}[htbp]
\centering
\caption{Calculation Methods of Stability and Uncertainty for PUCT Search.}
\label{tab:stability_computation}
\resizebox{\textwidth}{!}{
\begin{tabular}{@{}llll@{}}
\toprule
\textbf{Environment} & \textbf{Methodology} & \textbf{Action Space} & \textbf{Perturbation / Sampling Details} \\
\midrule

\multicolumn{4}{l}{\textit{\textbf{Strategy 1: Action Space Perturbation for Stability Estimation}}} \\
\multicolumn{4}{l}{\textit{Stability Score: \quad $\hat{p}_{\mathrm{stab}}(a)  = \frac{1}{J}\sum_{j=1}^{J} \hat{p}_{\mathrm{succ}}(o, a'_j)$}} \\
\cmidrule(l){1-4}

\textit{\textbf{PHYRE}} & \multirow{3}{*}{\begin{tabular}[c]{@{}l@{}}Action Space\\ Perturbation\end{tabular}} & $(x, y, r)$ Grid coord. \& radius & \begin{tabular}[c]{@{}l@{}}Perturb $(x,y)$ by $\pm1$ (4 directions), \\ and $r$ by $\pm1$ or $0$. \end{tabular} \\
\cmidrule(l){1-1} \cmidrule(l){3-4}

\textit{\textbf{I-PHYRE}} &  & \begin{tabular}[c]{@{}l@{}}Sequence of timed events\\ $[(i_k, t_k)]$\end{tabular} & \begin{tabular}[c]{@{}l@{}}Temporal jittering: perturb timestamps $t_k$ \\ with $\pm\Delta t \in \{\pm0.5\text{s}, \pm1.0\text{s}\}$ for a subset of events.\end{tabular} \\
\cmidrule(l){1-1} \cmidrule(l){3-4}

\textit{\angrybirdlogo} &  & $(\theta, p)$ Angle \& power & \begin{tabular}[c]{@{}l@{}}Perturb $\theta$ with $\Delta_\theta \in \{-5^\circ, 0^\circ, 5^\circ\}$ and \\ $p$ with $\Delta_p \in \{-0.1, 0, 0.1\}$.\end{tabular} \\ 
\midrule

\multicolumn{4}{l}{\textit{\textbf{Strategy 2: Confidence-Based Scoring for Highly Sensitive Environments}}} \\
\multicolumn{4}{l}{\textit{Uncertainty Score (LCB): \quad $\text{score}(a|o) = \mu_p - \lambda_{\mathrm{LCB}} \sigma_p$}} \\
\cmidrule(l){1-4}

\textit{\textbf{Kinetix}} & \multirow{3}{*}{\begin{tabular}[c]{@{}l@{}}Model Confidence\\ (LCB)\end{tabular}} & \multirow{3}{*}{\begin{tabular}[c]{@{}l@{}}Varies (\eg timings, \\ forces, positions)\end{tabular}} & \multirow{3}{*}{\begin{tabular}[c]{@{}l@{}}$K=8$ stochastic forward passes of $\mathcal{M}_\phi(o,a)$ \\ using different temperatures sampled from $[0.1, 1.0]$. \\ This yields a set of probabilities $\{p^{(j)}\}_{j=1}^K$.\end{tabular}} \\
\cmidrule(l){1-1} 
\textit{\pooltoollogo} & & & \\
\cmidrule(l){1-1} 
\textit{\cutrepologo} & & & \\
\bottomrule
\end{tabular}
}
\end{table}

\textit{Strategy 1: Action Space Perturbation.} For environments with stable physics (\textit{PHYRE}, \textit{I-PHYRE}, \textit{Angry Birds}), we estimate stability by sampling a neighborhood of $J$ perturbed actions $\{a'_j\}_{j=1}^J$. The stability score is the average success probability of these neighbors:
$
    \hat{p}_{\mathrm{stab}}(o,a) = \frac{1}{J} \sum_{j=1}^{J} \hat{p}_{\mathrm{succ}}(o, a'_j).
$
Specifically:
\begin{itemize}
    \item \textit{PHYRE}: We perturb grid coordinates $(x,y)$ by $\pm 1$ and radius $r$ by $\pm 1$ ($J=12$).
    \item \textit{I-PHYRE}: We apply temporal jittering $\delta \in \{\pm 0.5\text{s}, \pm 1.0\text{s}\}$ to event timestamps, ensuring causal order is preserved.
    \item \textit{Angry Birds}: We perturb launch angle $\theta$ by $\pm 5^\circ$ and power $p$ by $\pm 0.1$.
\end{itemize}
We combine the mean success prediction $\mu_p$ and this stability score into:
$
    \text{score}(a\mid o)=(1-\lambda_{\text{PUCT}})\,\mu_p(o,a)+\lambda_{\text{PUCT}}\,\hat{p}_{\mathrm{stab}}(o,a).
$

\textit{Strategy 2: Model Confidence (LCB).} For environments sensitive to initial conditions (\textit{Kinetix}, \textit{Pooltool}, \textit{Cut the Rope}), we instead estimate the model's epistemic uncertainty. We perform $K=8$ stochastic forward passes with temperatures sampled from $[0.1, 1.0]$ to obtain success probabilities $\{p^{(j)}\}_{j=1}^K$. We then compute a Lower Confidence Bound (LCB) score:
$
    \text{score}(a | o) = \mu_p - \lambda_{\mathrm{LCB}} \cdot \sigma_p,
$
where $\mu_p$ and $\sigma_p$ are the mean and standard deviation of predictions, and $\lambda_{\mathrm{LCB}}=0.2$. This penalizes uncertain actions in more chaotic domains.

\paragraph{PUCT Search.} These scores guide a search at the root node. For a planning budget of $B$ iterations, we select the action $a_t$ that maximizes the PUCT criterion:
\begin{align}
    a_t=\arg\max_{a\in A}\left[Q(a)+c_{\mathrm{PUCT}}\cdot P(a)\cdot \frac{\sqrt{\sum_{b \in A} N(b)}}{1+N(a)}\right]
\end{align}

Upon selecting $a_t$, its world model score is used to update $Q(a_t)$ and $N(a_t)$. We employ an \textbf{early-stopping} mechanism if the budget $B$ is exhausted, the same action is chosen 3 times consecutively, or a highly successful action ($\mu_p > 0.8$) is found.
Note that unlike standard MCTS where $Q(a)$ is updated via actual environment rollouts, here $Q(a)$ is maintained as a running average of world model scores $v$, serving as a proxy value estimate within the root-node planning budget.

\paragraph{Execution.} Finally, the action $a^* = \arg\max_a Q(a)$ is executed for a single step.

\section{Experiments}

\subsection{Experimental Setup}
\label{sec:experimental_setup}

This subsection details the experimental environments, comparison baselines, evaluation metrics, and the implementation details of \ProjName.

\subsubsection{Environments \& Task Specifications}
\label{sec:exp_setup_env}

We adopt DeepPHY~\citep{xu2025deepphy} as our evaluation platform. This benchmark comprises six diverse and interactive physical simulation environments: \textit{\textbf{PHYRE}} \citep{bakhtin2019phyre}, \textit{\textbf{I-PHYRE}} \citep{iphyre}, \textit{\textbf{Kinetix}} \citep{Kinetix}, \textit{\pooltoollogo} \citep{kiefl2024pooltool}, \textit{\angrybirdlogo}\footnote{https://apps.apple.com/us/app/rovio-classics-angry-birds/id1596736236}, and \textit{\cutrepologo}\footnote{https://apps.apple.com/cn/app/cut-the-rope/id1024507512}. Encompassing a wide spectrum of physical properties (\eg gravity, elasticity, collisions), these environments present challenges ranging from single-step planning (\textit{PHYRE}) to complex multi-step sequential control (\textit{Kinetix}, \textit{Cut the Rope}).

To ensure a rigorous interactive evaluation, we formally define the task structure and success metrics below:

\paragraph{Formal Definitions.}
\begin{itemize}
    \item \textbf{Episode (Task Instance):} An episode refers to the complete problem-solving process for a single specific level or puzzle configuration (\eg PHYRE template 00002:001). An episode concludes when the agent successfully solves the task or exhausts the maximum allowed attempts.
    
    \item \textbf{Attempt (Trial):} An episode consists of a sequence of up to $K$ attempts. At the start of attempt $k$, the agent receives the visual observation and the text history of previous failed trajectories $H_{k-1} = \{\tau_1, ..., \tau_{k-1}\}$ to perform in-context learning. For PHYRE and I-PHYRE (training environments), the limit is $K=10$. For evaluation environments like Pooltool and Angry Birds, limits are set according to Table~\ref{tab:benchmark_specs_part2}.
    
    \item \textbf{Action Horizon (Per Attempt):} This defines the temporal complexity of a single plan within one attempt.
    \begin{itemize}
        \item \textit{Single-step / In-advance:} The agent outputs a complete static plan (\eg placing one ball in PHYRE) or a sequence of timed actions (I-PHYRE) at the start. The simulator then executes the physics until stability.
        \item \textit{Sequential / On-the-fly:} The agent interacts with the environment in a turn-based manner (\eg Kinetix, Angry Birds), observing intermediate states between actions.
    \end{itemize}
\end{itemize}

\paragraph{Environment Specifications.}
We provide a detailed summary of the input modalities, action spaces, and success criteria for all environments in Table~\ref{tab:benchmark_specs_part1} and Table~\ref{tab:benchmark_specs_part2}.

\begin{table}[h]
\caption{Detailed Specifications of DeepPHY Environments (Part I): Input Modalities and Action Spaces.}
\label{tab:benchmark_specs_part1}
\centering
\resizebox{\textwidth}{!}{
\begin{tabular}{@{}l|ll@{}}
\toprule
\textbf{Environment} & \textbf{Input Modality} & \textbf{Action Space Format} \\ \midrule
\textbf{PHYRE} & Image + 8$\times$8 Grid Overlay & Discretized Selection: \texttt{Cell: [1-64], Radius: [1-8]} \\
\textbf{I-PHYRE} & Image + Index IDs & JSON Sequence: \texttt{[\{"time": t, "index": i\}, ...]} \\
\textbf{Kinetix} & Image + Motor/Thruster IDs & JSON Vector: \texttt{[m\_1, m\_2, ..., t\_1, t\_2]} \\
\textbf{\pooltoollogo} & 2D Top-down View & Discretized Selection: \texttt{Speed: [Low/Med/High], Strikespot: [Spin]} \\
\textbf{\angrybirdlogo} & Screenshot (Annotated) & Code: \texttt{[shoot(angle=int, power=float)]} \\
\textbf{\cutrepologo} & Screenshot (Annotated) & Code: \texttt{[cut\_pin(id)], [pop\_bubble(id)]}, etc. \\ \bottomrule
\end{tabular}
}
\end{table}

\begin{table}[h]
\caption{Detailed Specifications of DeepPHY Environments (Part II): Horizons and Success Criteria.}
\label{tab:benchmark_specs_part2}
\centering
\resizebox{0.75\textwidth}{!}{
\begin{tabular}{@{}l|ll@{}}
\toprule
\textbf{Environment} & \textbf{Horizon} & \textbf{Success Criteria ($r_{GT}=1$)} \\ \midrule
\textbf{PHYRE} & Single-step & Green ball touches Target. \\
\textbf{I-PHYRE} & Sequence & All Red balls fall into abyss. \\
\textbf{Kinetix} & Multi-step (Max 16) & Green shape touches Blue, avoids Red. \\
\textbf{\pooltoollogo} & Multi-step (Max 15) & Pot the 9-ball legally. \\
\textbf{\angrybirdlogo} & Multi-step (Birds avail.) & All pigs are eliminated. \\
\textbf{\cutrepologo} & Multi-step & Candy reaches \textit{Om Nom}'s mouth. \\ \bottomrule
\end{tabular}
}
\end{table}

\begin{figure*}[t]
    \centering
    \begin{subfigure}[b]{0.19\textwidth}
        \centering
        \includegraphics[trim={0cm 0cm 0cm 1cm}, clip, width=\linewidth]{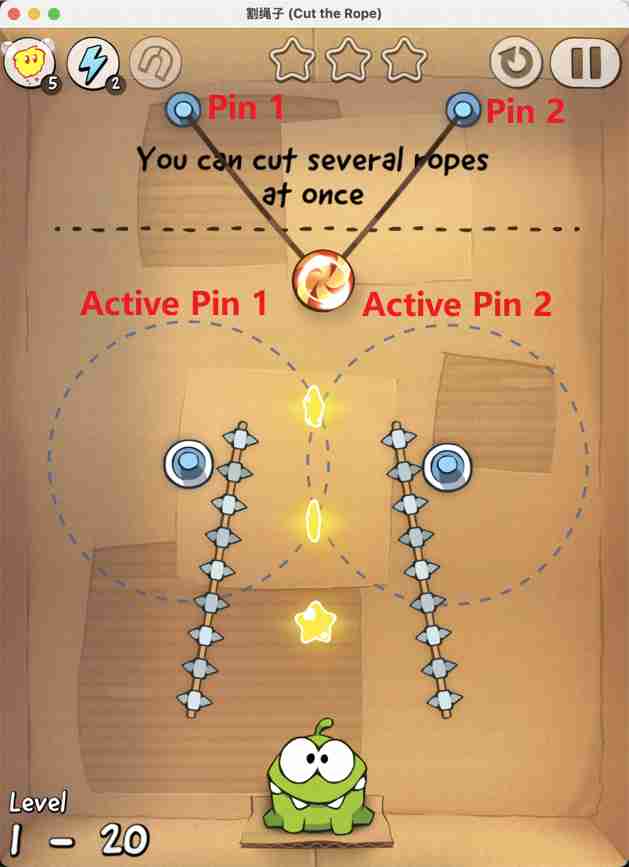}
    \end{subfigure}
    \hfill
    \begin{subfigure}[b]{0.19\textwidth}
        \centering
        \includegraphics[trim={0cm 0cm 0cm 1cm}, clip, width=\linewidth]{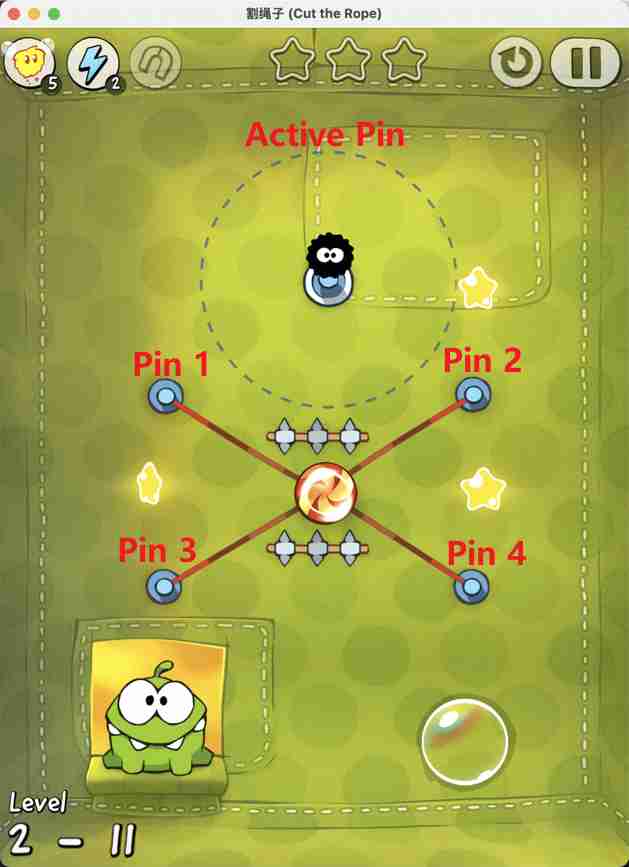}
    \end{subfigure}
    \hfill
    \begin{subfigure}[b]{0.19\textwidth}
        \centering
        \includegraphics[trim={0cm 0cm 0cm 1cm}, clip, width=\linewidth]{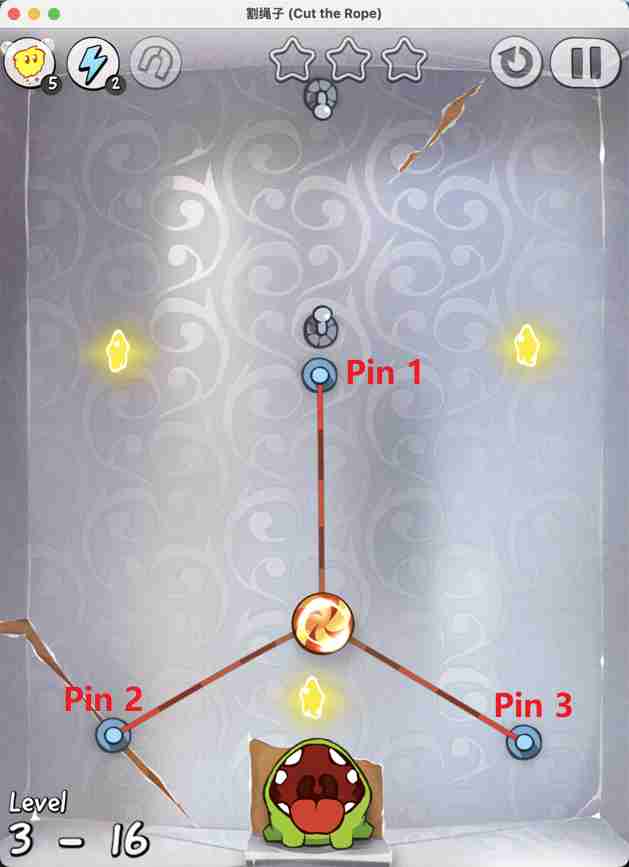}
    \end{subfigure}
    \hfill
    \begin{subfigure}[b]{0.19\textwidth}
        \centering
        \includegraphics[trim={0cm 0cm 0cm 1cm}, clip, width=\linewidth]{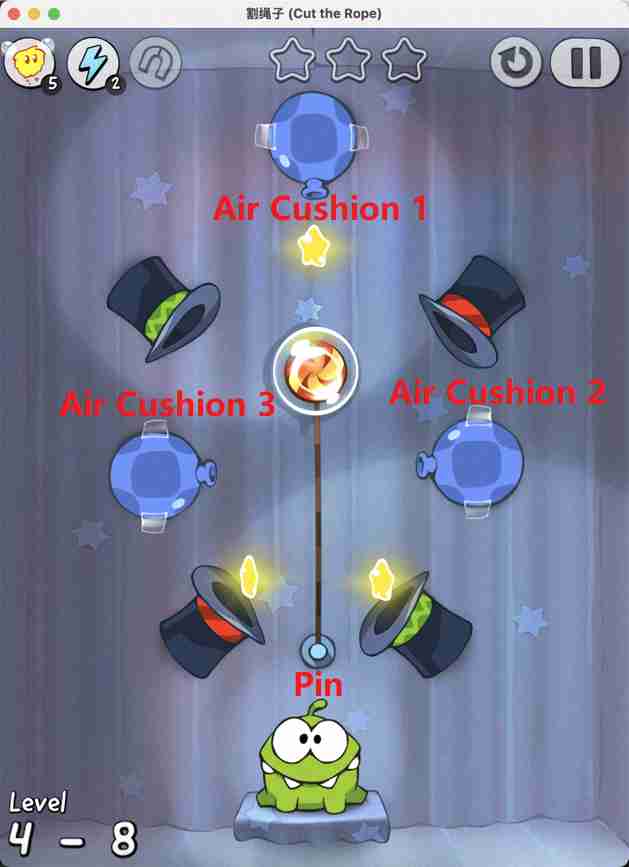}
    \end{subfigure}
    \hfill
    \begin{subfigure}[b]{0.19\textwidth}
        \centering
        \includegraphics[trim={0cm 0cm 0cm 1cm}, clip, width=\linewidth]{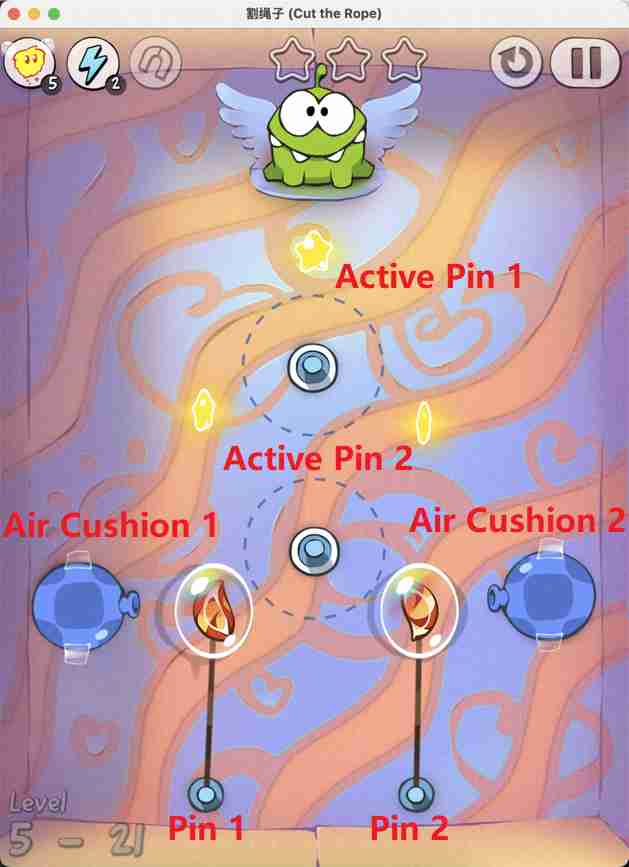}
    \end{subfigure}
    
    \caption{Examples of static element annotation in the \textit{Cut the Rope} game. Key static props—such as Pins, Active Pins, and Air Cushions—are clearly marked with numerical IDs. This method converts pixel-level visual information into grounded tokens, enabling the Agent to accurately identify and manipulate objects.}
\label{fig:cutrope_static_annotated}
\end{figure*}

\paragraph{Implementation Example: Cut the Rope.}
To illustrate how continuous gaming environments are adapted for VLM control, we detail the implementation of the \textit{Cut the Rope} task (visualized in Figure~\ref{fig:cutrope_static_annotated}).

\textit{Visual Input:} The model receives a screenshot where interactive elements are annotated with numerical IDs. For example, a rope anchored to a pin might be labeled ``Pin 1'', and a floating bubble containing candy might be labeled ``Bubble 2''.

\textit{Action Space:} Instead of continuous gestures, the model outputs structured Python-like function calls within square brackets. The interpreter parses these tokens into game actions. Supported commands include: \texttt{[cut\_pin(id=3)]} to cut a specific rope, \texttt{[pop\_bubble(id=2)]} to release candy, or \texttt{[sleep(seconds=0.5)]} to handle swing dynamics.

\subsubsection{Evaluation Protocol}

{To rigorously assess the acquisition of transferable physical intuition, we strictly define our evaluation protocol by partitioning the DeepPHY suite into source and target domains.}

{We train the Policy Model ($\pi_\theta$) and World Model ($\mathcal{M}_\phi$) exclusively on \textit{PHYRE} and \textit{I-PHYRE}. The models are evaluated on \textit{Kinetix}, \textit{Pooltool}, \textit{Angry Birds}, and \textit{Cut the Rope}. We enforce a strict zero-shot inference setting on the Target Domains. No weight updates, fine-tuning, or parameter calibration are permitted on these environments. The agent must rely solely on its learned physical intuition and in-context adaptation capabilities to solve these unseen tasks.}

\subsubsection{Baselines}

We compare \ProjName with leading closed-source and open-source VLMs, as detailed in Table~\ref{tab:model_abbreviations}. These closed-source models represent the current SOTA in zero-shot physical reasoning.
We use the random action agent (\texttt{\textit{MOCK}}) (provided by DeepPHY) as a lower bound on performance.

We evaluate two distinct variants of our own method: i) \textbf{\textsc{\ProjName} (Policy Only):} This variant performs inference using only our policy model, $\pi_\theta$, trained with GRPO, without leveraging the world model or PUCT search. This baseline is designed to measure the performance of the adaptive policy itself. And ii) \textbf{\textsc{\ProjName} (Full):} Our complete model, which combines the adaptive policy $\pi_\theta$ with the PUCT search guided by World Model $\mathcal{M}_\phi$, showcasing how both work in tandem.

\subsubsection{Evaluation Metrics}
We employ two core metrics for our evaluation: \textbf{Success Rate (S.R.)}, which is the percentage of tasks solved successfully; and \textbf{Average Attempts} (Avg. Att.), the mean number of attempts taken, calculated only on successfully solved tasks. We report
Success Rate and Average Attempts over three runs under the different settings.

\subsubsection{Implementation Details}
\textbf{Model Architecture:} Both Policy Model $\pi_\theta$ and World Model $\mathcal{M}_\phi$ are built and trained based on Qwen2.5-VL-3B/7B-Instruct~\citep{qwen2.5-VL}.

\textbf{Training:}
We conducted training for Policy Model $\pi_\theta$ and World Model $\mathcal{M}_\phi$ on the first two environments (\textit{PHYRE} \& \textit{I-PHYRE}\footnote{Although similarly named, the environments pose distinct challenges: \textit{PHYRE} centers on single-step actions that trigger complex chain reactions, while \textit{I-PHYRE} requires multi-step planning with precise temporal and sequential control.}), and then evaluated their generalization performance on the remaining four.
Policy Model $\pi_\theta$ is trained online using GRPO \citep{shao2024grpo}. For training, we set the policy learning rate to $1 \times 10^{-6}$. The policy updates are regularized with a KL penalty coefficient ($\beta$) of $0.001$. We employ a Turn-Aware Group Advantage Estimation scheme with a turn-level discount factor $\gamma_{\text{turn}}=0.95$ and a token-level discount $\gamma_{\text{token}}=1.0$. During the online data collection (rollout) phase, actions are sampled using a temperature of 0.7 and top-p nucleus sampling of 0.95.
World Model $\mathcal{M}_\phi$ is trained offline on the curated dataset, with the loss weight for the text component, $\lambda_{\text{text}}$, set to 0.2.

\textbf{Inference:}
The PUCT search is configured with a candidate sample size of $S=32$, a planning budget of $B=32$, a score mixing weight of $\lambda_{\text{PUCT}}=0.25$, and an exploration coefficient of $c_{\text{PUCT}}=1.5$. For environments using Strategy 1 (Stability), we use $J=12$ perturbed neighbors. For environments using Strategy 2 (LCB), we use $K=8$ stochastic forward passes with $\lambda_{\text{LCB}}=0.2$.

\subsection{Main Results}

\definecolor{seenheader}{HTML}{E1F5FE} 
\definecolor{unseenheader}{HTML}{FFF3E0}

\begin{table*}[h]
\centering
\caption{Overall Performance on DeepPHY. Model naming conventions are detailed in Appendix.}
\label{tab:overall_performance_deepphy}
\resizebox{\textwidth}{!}{
\begin{tabular}{l | ll  !{\vrule width 1.5pt} ll  ll !{\vrule width 1.5pt} l  ll l l}
\toprule
 
\multicolumn{3}{c!{\vrule width 1.5pt}}{\textbf{Model Configuration}}& 
\multicolumn{4}{c!{\vrule width 1.5pt}}{\cellcolor{seenheader}\textbf{In-Domain (Training)}} & 
\multicolumn{5}{c}{\cellcolor{unseenheader}\textbf{Out-of-Distribution (Zero-shot)}} \\ \midrule

\multirow{2}{*}{\textbf{\shortstack{Row \\ Ann.}}} & \multirow{2}{*}{\textbf{\shortstack{
\textcolor{policycolor}{Policy Model} \\ 
\textcolor{policycolor}{$\pi_\theta$}}}} 
& 
\multirow{2}{*}{\textbf{\shortstack{
\textcolor{worldcolor}{World Model} \\ \textcolor{worldcolor}{$\mathcal{M}_\phi$}
}}} & \multicolumn{2}{c}{\textbf{PHYRE}} & \multicolumn{2}{c!{\vrule width 1.5pt}}{\textbf{I-PHYRE}} & \multirow{2}{*}{\textbf{Kinetix}} & \multicolumn{2}{c}{\pooltoollogo} & \multirow{2}{*}{\angrybirdlogo} & \multirow{2}{*}{\cutrepologo} \\
\cmidrule(lr){4-5} \cmidrule(lr){6-7} \cmidrule(lr){9-10}
& & & \textbf{Att. 1} & \textbf{Att. 10} & \textbf{Att. 1} & \textbf{Att. 10} & & \textbf{Att. 1} & \textbf{Att. 15} &  &  \\
\midrule 
\textbf{\#\ 01.} & \texttt{\textit{MOCK}} & - & 1.50\% & 10.80\% & 23.33\%  &  53.33\% & 21.40\% & 2.33\%  & 48.00\% & 17.65\% & 11.36\% \\
\midrule
\rowcolor{lightgray}\multicolumn{12}{c}{\textbf{Open-Source Models}} \\ 
\midrule
\textbf{\#\ 02.} & \textbf{Qwen-3B} & - & 0.17\% & 5.85\% & 13.33\% & 16.67\%  & 16.22\% & 0.00\% & 50.00\% & 17.65\% & 7.95\%  \\
\textbf{\#\ 03.} & \textbf{Qwen-7B} & - & 0.67\% & 10.10\% & 32.42\%  & 32.42\%  & 13.51\% & 23.50\% & 26.50\% & 20.59\% & 9.09\%  \\
\textbf{\#\ 04.} & \textbf{Qwen-32B} & - & 0.03\% & 8.70\% & 0.17\% & 5.60\%  & 15.20\% & 0.00\% & 14.29\% & 26.47\% & 6.82\%  \\
\textbf{\#\ 05.} & \textbf{Qwen-72B} & - & 2.43\% & 14.92\% &  13.33\% & 43.33\%  & 14.86\% & 0.00\% & 18.00\% & 29.41\% & 13.64\%  \\
\textbf{\#\ 06.} & \textbf{Qwen-72B} & \textbf{Qwen-72B} & 1.78\% & 10.39\% & 10.00\% & 36.67\% & 12.16\% & 0.00\% & 14.00\% & 26.47\% & 11.50\% \\
\midrule
\rowcolor{lightgray}\multicolumn{12}{c}{\textbf{Close-Source Models}} \\
\midrule
\textbf{\#\ 07.} & \textbf{Claude 4.0 Opus} & - & 1.73\% & 10.63\% & 36.67\% &  56.67\% & 23.20\% & 0.00\% & 49.00\% & 32.35\% & {26.14\%} \firstlogo \\
\textbf{\#\ 08.} & \textbf{Claude 4.0 Opus} & \textbf{Claude 4.0 Opus} & 1.11\% & 7.44\% & 30.00\% & 50.00\% & 20.50\% & 0.00\% & 42.00\% & 29.41\% & {23.86\%} \secondlogo \\
\textbf{\#\ 09.} & \textbf{Gemini-2.5-Pro} & - & 2.17\% & 22.07\% & 20.00\% & 63.33\% & 24.10\% & 36.50\% & 68.00\% & 35.29\% & 22.73\% \\
\textbf{\#\ 10.} & \textbf{Gemini-2.5-Pro} & \textbf{Gemini-2.5-Pro} & 2.17\% & 12.33\% & 16.67\% & 53.33\% & 21.80\% & 25.00\% & 60.00\% & 29.41\% & 20.45\% \\
\textbf{\#\ 11.} & \textbf{GPT-o3} & - & 3.03\% & 30.77\% & 30.00\% & 86.67\% & {26.89\%} \firstlogo & 0.00\% & 25.67\% & 35.29\% & 18.18\% \\
\textbf{\#\ 12.} & \textbf{GPT-o3} & \textbf{GPT-o3} & 0.17\% & 25.60\% & 25.00\% & 76.67\% & {24.50\%} \secondlogo & 0.00\% & 22.00\% & 29.41\% & 16.50\% \\
\midrule
\rowcolor{lightgray}\multicolumn{12}{c}{\textbf{Fine-tuned Models}} \\ 
\midrule
\rowcolor{mylightgray}\multicolumn{12}{c}{\textbf{Qwen2.5-VL-3B-Instruct Fine-tuned Series}} \\
\textbf{\#\ 13.} & \textcolor{policycolor}{\textbf{Qwen-3B$^{\text{P}\ SFT}$}} & - & 1.33\% & 25.58\% & 20.00\% & 30.00\% & 13.70\% & 0.00\% & 51.00\% & 23.53\% & 10.23\% \\
\textbf{\#\ 14.} & \textcolor{policycolor}{\textbf{Qwen-3B$^{\text{P}\ SFT}$}} & \textcolor{worldcolor}{\textbf{Qwen-3B$^{\text{P}}$}} & 1.52\% & 28.69\% & 19.87\% & 32.45\% & 13.55\% & 0.00\% & 53.67\% & 26.47\% & 11.85\% \\

\textbf{\#\ 15.} & \textcolor{policycolor}{\textbf{Qwen-3B$^{\text{I}\ SFT}$}} & - & 0.33\% & 9.83\% & 23.33\% & 45.56\% & 9.60\% & 0.00\% & 43.00\% & 20.59\% & 7.80\% \\
\textbf{\#\ 16.} & \textcolor{policycolor}{\textbf{Qwen-3B$^{\text{I}\ SFT}$}} & \textcolor{worldcolor}{\textbf{Qwen-3B$^{\text{I}}$}} & 0.74\% & 11.88\% & 23.21\% & 47.34\% & 9.82\% & 0.00\% & 45.33\% & 23.53\% & 9.13\% \\

\textbf{\#\ 17.} & \textcolor{policycolor}{\textbf{Qwen-3B$^{\text{P\&I}\ SFT}$}} & - & 10.42\% & 29.00\% & 20.00\% & 50.00\% & 15.07\% & 0.00\% & 57.00\% & 26.47\% & 11.60\% \\
\textbf{\#\ 18.} & \textcolor{policycolor}{\textbf{Qwen-3B$^{\text{P\&I}\ SFT}$}} & \textcolor{worldcolor}{\textbf{Qwen-3B$^{\text{P\&I}}$}} & 10.70\% & 33.56\% & 21.29\% & 52.91\% & 15.11\% & 0.00\% & 59.67\% & 32.35\% & 14.05\% \\

\textbf{\#\ 19.} & \textcolor{policycolor}{\textbf{Qwen-3B$^{\text{P}\ GRPO}$}} & - & 9.50\% & 29.03\% & 13.33\% & 26.67\% & 12.33\% & 0.00\% & 58.00\% & 26.47\% & 11.05\% \\
\textbf{\#\ 20.} & \textcolor{policycolor}{\textbf{Qwen-3B$^{\text{P}\ GRPO}$}} & \textcolor{worldcolor}{\textbf{Qwen-3B$^{\text{P}}$}} & 9.85\% & 32.00\% & 12.78\% & 28.00\% & 12.33\% & 0.00\% & 61.00\% & 29.41\% & 12.98\% \\

\textbf{\#\ 21.} & \textcolor{policycolor}{\textbf{Qwen-3B$^{\text{I}\ GRPO}$}} & - & 0.17\% & 9.67\% & 36.67\% & 48.89\% & 8.20\% & 0.00\% & 50.00\% & 20.59\% & 7.75\% \\
\textbf{\#\ 22.} & \textcolor{policycolor}{\textbf{Qwen-3B$^{\text{I}\ GRPO}$}} & \textcolor{worldcolor}{\textbf{Qwen-3B$^{\text{I}}$}} & 0.66\% & 13.48\% & 37.60\% & 51.00\% & 8.07\% & 0.00\% & 52.33\% & 23.53\% & 9.88\% \\

\textbf{\#\ 23.} & \textcolor{policycolor}{\textbf{Qwen-3B$^{\text{P\&I}\ GRPO}$}} & - & 14.00\% & 34.50\% & 36.67\% & 60.00\% & 19.18\% & 0.00\% & 55.00\% & 29.41\% & 12.10\% \\
\textbf{\#\ 24.} & \textcolor{policycolor}{\textbf{Qwen-3B$^{\text{P\&I}\ GRPO}$}} & \textcolor{worldcolor}{\textbf{Qwen-3B$^{\text{P\&I}}$}} & 14.11\% & 39.49\% & 35.66\% & 61.99\% & 19.45\% & 0.00\% & 57.33\% & 35.29\% & 15.80\% \\
\midrule
\rowcolor{mylightgray}\multicolumn{12}{c}{\textbf{Qwen2.5-VL-7B-Instruct Fine-tuned Series}} \\

\textbf{\#\ 25.} & \textcolor{policycolor}{\textbf{Qwen-7B$^{\text{P}\ SFT}$}} & - & 3.67\% & 36.13\% & 13.33\% & 19.33\% & 15.07\% & 2.00\% & 66.00\% & 38.24\% & 15.35\% \\
\textbf{\#\ 26.} & \textcolor{policycolor}{\textbf{Qwen-7B$^{\text{P}\ SFT}$}} & \textcolor{worldcolor}{\textbf{Qwen-7B$^{\text{P}}$}} & 3.40\% & 38.67\% & 13.57\% & 22.15\% & 15.31\% & 2.00\% & 67.33\% & 41.18\% & 17.50\% \\

\textbf{\#\ 27.} & \textcolor{policycolor}{\textbf{Qwen-7B$^{\text{I}\ SFT}$}} & - & 0.67\% & 9.33\% & 12.22\% & 50.00\% & 13.70\% & 1.00\% & 64.00\% & 23.53\% & 8.21\% \\
\textbf{\#\ 28.} & \textcolor{policycolor}{\textbf{Qwen-7B$^{\text{I}\ SFT}$}} & \textcolor{worldcolor}{\textbf{Qwen-7B$^{\text{I}}$}} & 0.82\% & 13.45\% & 11.34\% & 51.50\% & 13.49\% & 1.00\% & 66.67\% & 26.47\% & 10.36\% \\

\textbf{\#\ 29.} & \textcolor{policycolor}{\textbf{Qwen-7B$^{\text{P\&I}\ SFT}$}} & - & 7.50\% & 42.67\% & 20.00\% & 63.33\% & 15.07\% & 2.00\% & 67.00\% & 41.18\% & 16.55\% \\
\textbf{\#\ 30.} & \textcolor{policycolor}{\textbf{Qwen-7B$^{\text{P\&I}\ SFT}$}} & \textcolor{worldcolor}{\textbf{Qwen-7B$^{\text{P\&I}}$}} & 7.95\% & {44.85\%} \secondlogo & 21.41\% & 65.38\% & 14.88\% & 2.00\% & {69.00\%} \secondlogo & {44.12\%} & 19.33\% \\

\textbf{\#\ 31.} & \textcolor{policycolor}{\textbf{Qwen-7B$^{\text{P}\ GRPO}$}} & - & 13.00\% & 40.00\% & 16.67\% & 28.89\% & 13.70\% & 0.00\% & 66.00\% & 41.18\% & 16.10\% \\
\textbf{\#\ 32.} & \textcolor{policycolor}{\textbf{Qwen-7B$^{\text{P}\ GRPO}$}} & \textcolor{worldcolor}{\textbf{Qwen-7B$^{\text{P}}$}} & 13.18\% & 43.66\% & 15.68\% & 30.00\% & 13.91\% & 0.00\% & {69.00\%} \secondlogo & {44.12\%} & 18.88\% \\

\textbf{\#\ 33.} & \textcolor{policycolor}{\textbf{Qwen-7B$^{\text{I}\ GRPO}$}} & - & 0.67\% & 8.33\% & 26.67\% & 86.67\% & 16.44\% & 0.00\% & 60.00\% & 23.53\% & 7.95\% \\
\textbf{\#\ 34.} & \textcolor{policycolor}{\textbf{Qwen-7B$^{\text{I}\ GRPO}$}} & \textcolor{worldcolor}{\textbf{Qwen-7B$^{\text{I}}$}} & 0.67\% & 13.04\% & 27.82\% & 89.31\% & 16.44\% & 0.00\% & 63.00\% & 26.47\% & 10.15\% \\
\textbf{\#\ 35.} & \textcolor{policycolor}{\textbf{Qwen-7B$^{\text{P\&I}\ GRPO}$}} & - & 14.63\% & 43.17\% & 13.33\% & {90.00\%} \secondlogo & 19.18\% & 0.00\% & {69.00\%} \secondlogo & {45.61\%} \secondlogo & 17.05\% \\ 
\textbf{\#\ 36.} & \textcolor{policycolor}{\textbf{Qwen-7B$^{\text{P\&I}\ GRPO}$}} & \textcolor{worldcolor}{\textbf{Qwen-7B$^{\text{P\&I}}$}} & 14.92\% & {45.56\%} \firstlogo & 13.12\% & {93.33\%} \firstlogo & 18.96\% & 0.00\% & {71.00\%} \firstlogo & {47.06\%} \firstlogo & 21.20\% \\
\bottomrule
\end{tabular}
}
\end{table*}

The main results, shown in Table \ref{tab:overall_performance_deepphy}, provide a comprehensive evaluation of the \textbf{\ProjName} framework and its components. Unlike the original DeepPHY benchmark, which evaluates on the complete task pool, we report \textbf{S.R.} on a held-out test split for \textit{PHYRE} and \textit{I-PHYRE}, as these  environments are used for training $\pi_\theta$ and  $\mathcal{M}_\phi$. \textbf{Crucially, all models in this  table were re-evaluated under this identical protocol}, ensuring a fair and consistent comparison. Their analysis validates our core hypotheses regarding in-context physical reinforcement learning, the role of world models, and the importance of training paradigms for generalization in complex physical reasoning tasks.

Our finetuned methods, including both SFT and GRPO, significantly improve the performance of the utilized base open-source VLM, Qwen2.5-VL. 
{The base models struggle to effectively solve these tasks, which demonstrates that our training pipeline successfully triggers and guides the VLMs to acquire physical intuition. Furthermore, after undergoing the identical fine-tuning process, the 7B models consistently outperform their 3B counterparts, highlighting the benefits of scaling model capacity.} 
The premier configuration, \textbf{\textcolor{policycolor}{Qwen-7B$^{\text{P\&I}\ \text{GRPO}}$}} paired with \textbf{\textcolor{worldcolor}{Qwen-7B$^{\text{P\&I}}$}} (Row \#36), consistently achieves state-of-the-art or comparable performance across nearly all tasks. This result validates our central hypothesis: the synergy between an adaptive policy ($\pi_\theta$), trained via GRPO on multi-episode interaction histories, and an independently-trained world model ($\mathcal{M}_\phi$), that guides planning, is critical for mastering complex physical challenges. Notably, this model not only excels in the trained environments (\textit{PHYRE} and \textit{I-PHYRE}) but also exhibits remarkable generalization to entirely unseen tasks in the other four diverse settings, demonstrating the acquisition of robust and transferable physical intuition.

A direct comparison between models trained with SFT (\eg Row \#29) and those trained with GRPO (\eg Row \#35) reveals the clear superiority of the latter. While SFT enables learning from expert trajectories, GRPO's online policy optimization teaches the agent to dynamically \textbf{adapt its strategy in-context} by conditioning on a history of successes and failures. This capability is fundamental to the ICRL paradigm, resulting in a more robust and adaptive policy that is particularly effective in tasks that inherently involve trial-and-error and iterative problem-solving.

The contribution of World Model ($\mathcal{M}_\phi$) is pivotal, transforming the agent from a purely reactive decision-maker into a deliberative planner. Across all finetuned pairs in the table, the full \ProjName configuration (Policy + World Model) consistently outperforms the policy-only variant. This confirms that the world model, acting as an in-context physical simulator, provides crucial foresight. The resulting ``propose-verify-select'' mechanism, where the policy generates candidate actions and the world model guides a PUCT search to select the most promising one, elevates the agent's reasoning from simple reaction to improved look-ahead planning.
Conversely, employing a generic, non-fine-tuned VLM as a world model (comparing pairs from Rows \#5 vs. \#6 to \#11 vs. \#12) degrades performance. This finding, consistent with the original DeepPHY, underscores that general-purpose VLMs currently lack fine-grained interactive physical reasoning capabilities.

Moreover, models trained jointly on both \textit{PHYRE} and \textit{I-PHYRE} (marked as \textbf{P\&I}) demonstrate superior performance on unseen tasks, when compared to models trained on either environment in isolation (\cf Rows \#31, \#33, and \#35). Specifically, the \textbf{\textcolor{policycolor}{Qwen-7B$^{\text{P\&I}\ \text{GRPO}}$}} policy (Row \#35) outperforms its single-environment counterparts across three unseen testbeds. This supports our hypothesis that exposure to diverse physical dynamics enables the model to internalize a more generalizable ``policy-improvement operator,'' facilitating the transfer of learned physical intuition to new domains.

In the \textit{Kinetix} environment, however, our models do not exhibit the same magnitude of performance gain as seen in the other environments. We attribute this to its unique nature, which demands precise fine-grained control of sub-object components and their tight interactions. In contrast, the other environments primarily involve reasoning about the holistic behavior and trajectory of whole objects.

\textit{Cut the Rope} presents a distinct challenge of its own: it combines GUI-based interaction (cutting ropes, popping bubbles, and timing actions) with complex multi-body physics simulation, making it arguably the most demanding environment in the benchmark.
Despite this, \ProjName achieves 21.20\% success rate---more than doubling the performance of the untrained Qwen-7B base model (9.09\%)---demonstrating that the physical intuition acquired during training does transfer meaningfully to this compound interactive domain, even under strict zero-shot conditions.

\subsection{Ablation Studies}
\label{subsec:ablation}

Our primary SFT models are trained on a dataset of successful solution trajectories from expert models (Gemini-2.5-Pro, GPT-4o, etc., detailed in Appendix), which often include multiple attempts (typically 5--10) before reaching a solution. This strategy is founded on the hypothesis that exposure to this trial-and-error process, even within a supervised framework, enables the model to internalize an iterative problem-solving strategy, aligning with the principles of ICRL.

To validate this hypothesis, we conduct a controlled ablation study detailed in Table~\ref{tab:ablation_single_vs_multi}. We compare our standard SFT models against ablated variants (denoted with a `$^{\text{single}}$' subscript). These ablated models are trained exclusively on a dataset of first-attempt successful trajectories generated via a systematic enumeration process. To ensure the comparison is fair, the number of training samples was kept identical for both SFT methods within the same environment.

\begin{table*}[h!]
\centering
\caption{\textbf{Ablation Study: }Performance of Multi-Attempt vs. Single-Attempt Training Trajectories.}
\label{tab:ablation_single_vs_multi}

\begin{subtable}{0.48\textwidth}
    \centering
    \caption{\textit{PHYRE} Benchmark.}
    \label{tab:model_performance_single_multi_phyre}
    \resizebox{\linewidth}{!}{
        \begin{tabular}{l | c | rrrrr}
        \toprule
        
        \textbf{Policy Model Only} & \textbf{Avg. Att.} & \textbf{Att. 1} & \textbf{Att. 4} & \textbf{Att. 7} & \textbf{Att. 10} \\
        \midrule
        
        \texttt{\textit{MOCK}} & 5.00 & 1.50\% & 5.87\% & 8.60\% & 10.80\% \\
        
        \midrule
        
        \textbf{Qwen-3B} & 3.98 & 0.17\% & 4.69\% & 5.67\% & 5.85\%  \\
        \textcolor{policycolor}{\textbf{Qwen-3B$^{\text{P}_{\text{single}}\ SFT}$}} & 2.60 & 7.63\% & 17.20\% & 19.83\% & 20.97\% \\
        \textcolor{policycolor}{\textbf{Qwen-3B$^{\text{P}\ SFT}$}}
        & 2.89 & 1.33\% & 16.58\% & 20.33\% & 25.58\% \\
        \textcolor{policycolor}{\textbf{Qwen-3B$^{\text{P}\ GRPO}$}} 
        & 3.23 & 9.50\% & 25.13\% & 28.80\% & 29.03\% \\
        
        \midrule
        
        \textbf{Qwen-7B} & 5.10 & 0.67\% & 5.33\% & 8.17\% & 10.10\% \\
        \textcolor{policycolor}{\textbf{Qwen-7B$^{\text{P}_{\text{single}}\ SFT}$}} & 2.61 & 13.17\% & 21.58\% & 26.97\% & 29.14\% \\
        \textcolor{policycolor}{\textbf{Qwen-7B$^{\text{P}\ SFT}$}} & 3.74 & 3.67\% & 27.33\% & 34.20\% & {36.13\%} & \secondlogo \\
        \textcolor{policycolor}{\textbf{Qwen-7B$^{\text{P}\ GRPO}$}} & 5.70 & 13.00\% & 33.17\% & 41.77\% & {40.00\%} & \firstlogo \\
        
        \midrule
        
        \textbf{Qwen-32B} & 3.97 & 0.03\% & 3.10\% & 6.93\% & 8.70\% \\
        \textbf{Qwen-72B} & 4.48 & 2.43\% & 9.53\% & 12.40\% & 14.92\% \\
        
        \bottomrule
        \end{tabular}
    }
\end{subtable}
\hfill
\begin{subtable}{0.48\textwidth}
    \centering
    \caption{\textit{I-PHYRE} Benchmark.}
    \label{tab:model_performance_single_multi_iphyre}
    \resizebox{\linewidth}{!}{
        \begin{tabular}{l | c | rrrrr}
        \toprule
        
        \textbf{Policy Model Only} & \textbf{Avg. Att.} & \textbf{Att. 1} & \textbf{Att. 4} & \textbf{Att. 7} & \textbf{Att. 10} \\
        \midrule
        
        \texttt{\textit{MOCK}} & 3.81 & 23.33\% & 43.33\% & 46.67\% & 53.33\% \\
        
        \midrule
        
        \textbf{Qwen-3B} & 2.87 & 13.33\% & 16.67\% & 16.67\% & 16.67\% \\
        \textcolor{policycolor}{\textbf{Qwen-3B$^{\text{I}_{\text{single}}\ SFT}$}} & 2.97 & 33.43\% & 33.43\% & 34.54\% & 34.54\% \\
        \textcolor{policycolor}{\textbf{Qwen-3B$^{\text{I}\ SFT}$}} & 3.72 & 23.33\% & 30.00\% & 43.33\% & 45.56\% \\
        \textcolor{policycolor}{\textbf{Qwen-3B$^{\text{I}\ GRPO}$}} & 3.81 & 36.67\% & 43.33\% & 45.56\% & 48.89\% \\
        
        \midrule
        
        \textbf{Qwen-7B} & 2.20 & 32.42\% & 32.42\% & 32.42\% & 32.42\% \\
        \textcolor{policycolor}{\textbf{Qwen-7B$^{\text{I}_{\text{single}}\ SFT}$}} & 3.44 & 35.56\% & 35.56\% & 35.56\% & 35.56\% \\
        \textcolor{policycolor}{\textbf{Qwen-7B$^{\text{I}\ SFT}$}} & 3.53 & 12.22\% & 46.67\% & 50.00\% & 50.00\% & \secondlogo \\
        \textcolor{policycolor}{\textbf{Qwen-7B$^{\text{I}\ GRPO}$}} & 4.37 & 26.67\% & 43.33\% & 53.33\% & 86.67\% & \firstlogo \\
        
        \midrule
        
        \textbf{Qwen-32B} & 1.40 & 0.17\% & 5.60\% & 5.60\% & 5.60\% \\
        \textbf{Qwen-72B} & 2.00 & 13.33\% & 30.00\% & 43.33\% & 43.33\% \\
        
        \bottomrule
        \end{tabular}
    }
\end{subtable}
\end{table*}

\begin{table*}[h!]
\centering
\caption{\textbf{Ablation Study: }Impact of Dynamic Visual Feedback on World Model Training.}
\label{tab:ablation_dynamic_frames}

\begin{subtable}{0.48\textwidth}
    \centering
    \caption{\textit{PHYRE} Benchmark.}
    \label{tab:ablation_wm_phyre}
    \resizebox{\linewidth}{!}{
        \begin{tabular}{l l| rr}
        \toprule
        
        \textbf{Policy Model} & \textbf{World Model} & \textbf{Att. 1} & \textbf{Att. 10} \\
        \midrule
        \textcolor{policycolor}{\textbf{Qwen-3B$^{\text{P}\ SFT}$}} & - & 1.33\% & 25.58\% \\
        \textcolor{policycolor}{\textbf{Qwen-3B$^{\text{P}\ SFT}$}} & \textcolor{worldcolor}{\textbf{Qwen-3B$^{\text{P}}_{w/o\ 5\ Frames}$}} & 1.40\% & 27.19\% \\
        \textcolor{policycolor}{\textbf{Qwen-3B$^{\text{P}\ SFT}$}} & \textcolor{worldcolor}{\textbf{Qwen-3B$^{\text{P}}$}} &  1.52\% & 28.69\%  \\

        \midrule
        \textcolor{policycolor}{\textbf{Qwen-3B$^{\text{P}\ GRPO}$}} & - & 9.50\% & 29.03\% \\
        \textcolor{policycolor}{\textbf{Qwen-3B$^{\text{P}\ GRPO}$}} & \textcolor{worldcolor}{\textbf{Qwen-3B$^{\text{P}}_{w/o\ 5\ Frames}$}} & 9.95\% & 30.20\% \\
        \textcolor{policycolor}{\textbf{Qwen-3B$^{\text{P}\ GRPO}$}} & \textcolor{worldcolor}{\textbf{Qwen-3B$^{\text{P}}$}} & 9.85\% & 32.00\% \\

        \midrule
        \textcolor{policycolor}{\textbf{Qwen-7B$^{\text{P}\ SFT}$}} & - &  3.67\% & 36.13\%  \\
        \textcolor{policycolor}{\textbf{Qwen-7B$^{\text{P}\ SFT}$}} & \textcolor{worldcolor}{\textbf{Qwen-7B$^{\text{P}}_{w/o\ 5\ Frames}$}} & 3.40\% &	37.47\% \\
        \textcolor{policycolor}{\textbf{Qwen-7B$^{\text{P}\ SFT}$}} & \textcolor{worldcolor}{\textbf{Qwen-7B$^{\text{P}}$}} & 3.40\% & 38.67\% \\

        \midrule
        \textcolor{policycolor}{\textbf{Qwen-7B$^{\text{P}\ GRPO}$}} & - & 13.00\% & 40.00\% \\
        \textcolor{policycolor}{\textbf{Qwen-7B$^{\text{P}\ GRPO}$}} & \textcolor{worldcolor}{\textbf{Qwen-7B$^{\text{P}}_{w/o\ 5\ Frames}$}} & 12.68\% & 41.76\%\\
        \textcolor{policycolor}{\textbf{Qwen-7B$^{\text{P}\ GRPO}$}} & \textcolor{worldcolor}{\textbf{Qwen-7B$^{\text{P}}$}} & 13.18\% & 43.66\% \\
        
        \bottomrule
        \end{tabular}
    }
\end{subtable}
\hfill
\begin{subtable}{0.48\textwidth}
    \centering
    \caption{\textit{I-PHYRE} Benchmark.}
    \label{tab:ablation_wm_iphyre}
    \resizebox{\linewidth}{!}{
        \begin{tabular}{l l | rr}
        \toprule
        
        \textbf{Policy Model} & \textbf{World Model} & \textbf{Att. 1} & \textbf{Att. 10} \\
        \midrule

        \textcolor{policycolor}{\textbf{Qwen-3B$^{\text{I}\ SFT}$}} & - & 23.33\% & 45.56\%  \\
        
        \textcolor{policycolor}{\textbf{Qwen-3B$^{\text{I}\ SFT}$}} & \textcolor{worldcolor}{\textbf{Qwen-3B$^{\text{I}}_{w/o\ 5\ Frames}$}} & 22.81\% & 45.04\% \\
        
        \textcolor{policycolor}{\textbf{Qwen-3B$^{\text{I}\ SFT}$}} & \textcolor{worldcolor}{\textbf{Qwen-3B$^{\text{I}}$}} & 23.21\% & 47.34\%  \\

        \midrule
        
        \textcolor{policycolor}{\textbf{Qwen-3B$^{\text{I}\ GRPO}$}} & - & 36.67\% & 48.89\% \\
        \textcolor{policycolor}{\textbf{Qwen-3B$^{\text{I}\ GRPO}$}} & \textcolor{worldcolor}{\textbf{Qwen-3B$^{\text{I}}_{w/o\ 5\ Frames}$}} & 36.97\% & 47.29\% \\
        \textcolor{policycolor}{\textbf{Qwen-3B$^{\text{I}\ GRPO}$}} & \textcolor{worldcolor}{\textbf{Qwen-3B$^{\text{I}}$}} & 37.60\% & 51.00\%  \\

        \midrule

        \textcolor{policycolor}{\textbf{Qwen-7B$^{\text{I}\ SFT}$}} & - & 12.22\% & 50.00\%  \\

        \textcolor{policycolor}{\textbf{Qwen-7B$^{\text{I}\ SFT}$}} & \textcolor{worldcolor}{\textbf{Qwen-7B$^{\text{I}}_{w/o\ 5\ Frames}$}} & 11.34\% & 50.30\% \\
        
        \textcolor{policycolor}{\textbf{Qwen-7B$^{\text{I}\ SFT}$}} & \textcolor{worldcolor}{\textbf{Qwen-7B$^{\text{I}}$}} & 11.34\% & 51.50\% \\

        \midrule

        \textcolor{policycolor}{\textbf{Qwen-7B$^{\text{I}\ GRPO}$}} & - & 26.67\% & 86.67\% \\
        \textcolor{policycolor}{\textbf{Qwen-7B$^{\text{I}\ GRPO}$}} & \textcolor{worldcolor}{\textbf{Qwen-7B$^{\text{I}}_{w/o\ 5\ Frames}$}} & 27.12\% & 87.51\% \\
        \textcolor{policycolor}{\textbf{Qwen-7B$^{\text{I}\ GRPO}$}} & \textcolor{worldcolor}{\textbf{Qwen-7B$^{\text{I}}$}} & 27.82\% & 89.31\% \\

        \bottomrule
        \end{tabular}
    }
\end{subtable}
\end{table*}

While the `$^{\text{single}}$' models demonstrate strong initial performance (Att. 1), they often plateau, a behavior particularly pronounced on the \textit{I-PHYRE} benchmark. In contrast, models trained on multi-attempt data exhibit a significantly steeper improvement curve, ultimately achieving superior performance by the final attempts (Att. 10). This confirms that learning from a history of failures and recoveries is crucial for developing a more robust policy. Furthermore, the success rate for our models trained on multi-attempt trajectories consistently and monotonically increases as the interaction history lengthens. A compelling trend is found where the average number of attempts for successful solutions (Avg. Att.) tends to increase in tandem with the model's capability (\eg from \textcolor{policycolor}{\textbf{SFT}} to \textcolor{policycolor}{\textbf{GRPO}} variants). This suggests that more advanced models are not simply more efficient, but are capable of tackling more complex problems that inherently require a longer iterative process. This provides direct empirical evidence that the models have learned to leverage ICL to interact with the environment and parse physical intuition. 
Finally, the outstanding performance of the \textcolor{policycolor}{\textbf{GRPO}} models further underscores the efficacy of our approach.

Table \ref{tab:ablation_dynamic_frames} evaluates the importance of providing the World Model with visual information about the dynamic consequences of an action during its training phase. We compare the full model against a variant where the World Model was trained without the five uniformly sampled post-action frames (detailed in Algorithm~\ref{alg:wm_data_curation}).
Across both \textit{PHYRE} and \textit{I-PHYRE}, and for all policy model configurations, the full \ProjName framework---which leverages a world model trained with post-action visual frames---consistently outperforms the variant employing an ablated world model (data curated by \texttt{w/o 5 Frames}). This performance delta provides strong evidence that incorporating explicit visual feedback of an action's dynamic consequences is crucial for training an effective world model. 
Which, in turn, enhances its predictive fidelity, directly translating to more effective guidance for the PUCT search procedure at inference time. Furthermore, it is noteworthy that even the ablated world model generally provides some performance lift over the policy-only baseline, underscoring the fundamental utility of our decoupled two-component architecture for deliberative planning.

\section{Conclusion}

In this work, we introduced \textbf{\ProjName} (In-Context Physical Reinforcement Learning), a framework designed to address the limitations of existing VLMs in interactive reasoning within dynamic physical environments. Our framework integrates {an} adaptive policy model with {a} world model that provides explicit physical intuition. 
Our extensive evaluations on the diverse DeepPHY benchmark demonstrate that \ProjName not only achieves significant performance gains over strong baselines but, crucially, maintains robust generalization {and} enables {the} genuine in-context acquisition of an environment's physical dynamics directly from interactive experience.

While \ProjName represents a significant step forward, this work also highlights promising avenues for future research. First, the policy and world models in the current framework are trained independently. Future work could explore synergistic training paradigms where the models are co-trained, allowing data generated by one to inform and enhance the learning process of the other, potentially creating a virtuous cycle of improvement. 
{Second, our analysis revealed that while \ProjName excels at learning overarching physical principles, its performance gains were less pronounced on tasks like \textit{Kinetix}, which demands high-dexterity, component-level manipulation. This distinction suggests that such fine-grained control problems may constitute a distinct class of challenges, representing another promising direction for future investigation.}

\clearpage
\bibliographystyle{unsrt}
\bibliography{ref}

\clearpage
\appendix
\section*{Appendix: Model Abbreviations}
\label{sec:abbreviations}

The abbreviations used throughout the paper and the full details of the corresponding open-source and closed-source models are listed in Table~\ref{tab:model_abbreviations}.

\begin{table}[h]
    \centering
    \caption{List of Model Abbreviations}
    \label{tab:model_abbreviations}
    \begin{tabular}{lll}
        \toprule
        \textbf{Model} & \textbf{Reference} & \textbf{Abbreviation} \\
        \midrule
        \texttt{\textit{MOCK}} & - & \texttt{\textit{MOCK}} \\
        \rowcolor{lightgray}\multicolumn{3}{c}{\textbf{Open-Source Models}} \\ \midrule
        \rowcolor{mylightgray}\multicolumn{3}{c}{\textbf{Qwen2.5-VL Series}} \\
        Qwen2.5-VL-3B-Instruct & - & Qwen-3B \\
        Qwen2.5-VL-7B-Instruct & - & Qwen-7B \\
        Qwen2.5-VL-32B-Instruct & - & Qwen-32B \\
        Qwen2.5-VL-72B-Instruct & - & Qwen-72B \\
        \bottomrule
        \rowcolor{lightgray}\multicolumn{3}{c}{\textbf{Closed-Source Models}} \\ \midrule
        Claude 4.0 Opus & \citep{anthropic2025claude} & Claude 4.0 Opus \\
        Gemini-2.5-Pro-06-17 & \citep{comanici2025gemini2.5} & Gemini-2.5-Pro \\
        GPT-o3-0416 & \citep{openai2025o3ando4mini} & GPT-o3 \\
        \bottomrule
    \end{tabular}
\end{table}

To clearly distinguish the fine-tuned models developed in our work, we adopt a systematic naming convention, which we illustrate using the Qwen-3B model as an example. This convention applies equally to Qwen-7B-based models.
We use \textcolor{policycolor}{\texttt{GREEN}} to denote \textcolor{policycolor}{Policy Models} and \textcolor{worldcolor}{\texttt{ORANGE}} for \textcolor{worldcolor}{World Models}.

\paragraph{Policy Models.} The nomenclature for our \textcolor{policycolor}{Policy Models} follows the format: 
\textcolor{policycolor}{BaseModel$^{\text{Env}_{\text{Data-Type}}\ Method}$}. Each component in the naming structure is defined as follows:

\begin{itemize}
    \item \textbf{Env:} Indicates the environment(s) used in fine-tuning:
    \begin{itemize}
        \item \textbf{P}: Denotes fine-tuning exclusively on \textit{PHYRE}.
        \item \textbf{I}: Denotes fine-tuning exclusively on \textit{I-PHYRE}.
        \item \textbf{P\&I}: Denotes a model trained using tasks from both \textit{PHYRE} and \textit{I-PHYRE}.
    \end{itemize}
    \item \textbf{Method:} Specifies the training algorithm used:
    \begin{itemize}
        \item \textbf{SFT}: Supervised Fine-Tuning.
        \item \textbf{GRPO}: Group Relative Policy Optimization \citep{shao2024grpo}.
    \end{itemize}
    \item \textbf{Data-Type:} An optional subscript specifying the nature of the training data:
    \begin{itemize}
        \item \textbf{(No subscript)}: The model was trained on a dataset of successful solution trajectories collected from expert models (\eg Gemini-2.5-Pro, GPT-4o). These trajectories were often generated over multiple attempts (typically 5--10) to find a solution. This strategy enables the model to learn from a process of trial and error, aligning with the philosophy of ICRL.
        \item \textbf{$^{\text{single}}$}: The model was trained exclusively on a dataset of first-attempt successful trajectories, generated via systematic enumeration over the action space.
    \end{itemize}
\end{itemize}

For example, \textcolor{policycolor}{\textbf{Qwen-3B$^{\text{P}_{\text{single}}\ SFT}$}} denotes a policy model based on Qwen2.5-VL-3B-Instruct, trained via SFT on single-attempt successful trajectories from the \textit{PHYRE} environment.

\paragraph{World Models.} The naming for our \textcolor{worldcolor}{World Models} is more concise. Since all World Models are trained via SFT, the naming directly specifies the base model and the environment. For example, \textcolor{worldcolor}{\textbf{Qwen-3B$^{\text{P}}$}} indicates the model is based on Qwen-3B and trained on \textit{PHYRE}.

World models may include one subscript: ``\textcolor{worldcolor}{\textbf{$_{w/o\ 5\ Frames}$}}". This subscript stands for ``without 5 frames'' and represents a critical ablation detail. As described in Subsection \ref{subsec:ablation}, this variant was trained on a curated dataset that \textit{excluded} the five uniformly sampled post-action video frames, serving as a baseline to validate the importance of dynamic visual feedback.

\end{document}